\documentclass[fullname]{clv2} 
\usepackage{graphicx}
\usepackage[space]{grffile}
\usepackage{latexsym}
\usepackage{amsfonts,amsmath,amssymb}
\usepackage{url}
\usepackage[utf8]{inputenc}
\usepackage{fancyref}
\usepackage{color}

\renewcommand{\footmark}{}

\dochead{}

\runningtitle{Computational Sociolinguistics: A Survey }

\runningauthor{Nguyen et al.}
\begin{document}

\title{Computational Sociolinguistics: A Survey}

\author{Dong Nguyen}
\affil{University of Twente}

\author{A. Seza Do\u{g}ru\"{o}z}
\affil{Tilburg University/\\Netherlands Institute for Advanced Study in the Humanities and Social Sciences (NIAS)}

\author{Carolyn P. Ros\'{e}}
\affil{Carnegie Mellon University}

\author{Franciska de Jong}
\affil{University of Twente/\\Erasmus University Rotterdam}

\maketitle

\begin{abstract}
Language is a social phenomenon and variation is inherent to its social nature. Recently, there has been a surge of interest within the computational linguistics (CL) community in the social dimension of language. In this article we present a survey of the emerging field of ‘Computational Sociolinguistics’ that reflects this increased interest. We aim to provide a comprehensive overview of CL research on sociolinguistic themes, featuring topics
such as the relation between language and social identity, language use in social interaction and  multilingual communication. Moreover, we demonstrate the potential for synergy between  the research communities involved, by showing how  the large-scale data-driven methods that are widely used in CL can complement existing sociolinguistic studies, and how sociolinguistics can inform and challenge the methods and assumptions
employed in CL studies. We hope to convey the possible benefits of a  closer collaboration between the two communities and conclude with a discussion of open challenges.
\end{abstract}

\section{Introduction}

Science has experienced a paradigm shift along with the increasing availability of large amounts of digital research data \cite{hey2009fourth}. In addition to the traditional focus on the description of natural phenomena, theory development and computational science, data-driven exploration and discovery have become a dominant ingredient of many methodological frameworks. In line with these developments, the field of computational linguistics (CL) has also evolved. 

Human communication occurs in both verbal and nonverbal form. Research
on computational linguistics has primarily focused on capturing the informational dimension of language 
and the structure of verbal information transfer. 
In the words of \namecite{lebowsky2015}, computational linguistics has made great progress in modeling language's informational dimension, but with a few notable exceptions, computation has had little to contribute to our understanding of language's social dimension.
The recent increase in interest of computational linguists to study language in social contexts is partly driven
by the ever increasing availability of social media data. 
Data from social media platforms provide a strong
incentive for innovation in the CL research agenda and  the surge
in relevant data  opens up methodological possibilities for
studying text as social data. Textual
resources, like many other language resources, can be seen as a data
type that is signaling all kinds of social phenomena.
This is related to the fact that language is one of the instruments
by which people construct their online identity and manage their
social network. There are challenges as well. For example, social media language is more colloquial and contains more linguistic variation, such as the use of slang and dialects, than the language in  datasets that have been commonly used in CL research (e.g., scientific articles, newswire text and the Wall Street Journal) \cite{eisenstein2013bad}. However, an even greater challenge is that the relation between social variables and language is typically fluid and tenuous, while the CL field commonly focuses on the level of literal meaning and language structure, which is more stable.
 
The tenuous connection between social variables and language arises
because of the symbolic nature of the relation between them.  
With the language chosen a social identity is signaled, which may buy a speaker\footnote{We use the term `speaker' for an individual who has produced a message,
either as spoken word or in textual format. When discussing particular social media sites,
we may refer to 'users' as well.} something in terms of footing within a
conversation, or in other words: for speakers there is room for choice in how to use their linguistic repertoire in order to achieve social goals. This  freedom of choice is
often referred to as the agency of speakers and the linguistic symbols chosen can be thought of as a form of social currency. Speakers may thus make use of specific words or
stylistic elements to represent themselves in a certain way. However,
because of this agency, social variables cease to have an essential
connection with language use. It may be the case, for example, that on
average female speakers display certain characteristics in their
language more frequently than their male counterparts. Nevertheless, in specific circumstances, females may choose to de-emphasize their identity as females by modulating their language usage to sound more male.  Thus, while this exception serves to highlight rather than challenge the commonly accepted symbolic association between gender and language, it nevertheless means that it is less feasible to predict how a female will sound in a randomly selected context.  

Speaker agency also enables creative violations of conventional language patterns.  Just as with any violation of expectations, these creative violations communicate indirect meanings. As these violations become conventionalized, they may be one vehicle towards language change.  Thus,  agency plays a role in explaining the variation in and dynamic nature of language practices, both within individual speakers and across speakers.  This variation is manifested at various levels of expression --
the choice of lexical elements, phonological variants, semantic
alternatives and grammatical patterns -- and plays a central role in the phenomenon of linguistic change. The audience, demographic
variables (e.g., gender, age), and speaker goals are among the
factors that influence how variation is exhibited in specific contexts. Agency thus increases the intricate complexity of language that must be captured in order to achieve a social interpretation of language.

Sociolinguistics investigates the reciprocal influence of society and language on
each other. Sociolinguists traditionally work with spoken data using qualitative and
quantitative approaches. Surveys and ethnographic research have been the main methods
of data collection \cite{eckert1989jocks,milroy1985linguistic,milroy2003,tagliamonte2006analysing,Trudgill1974,labov1968empirical}. The datasets used are often selected
and/or constructed to facilitate controlled statistical analyses and insightful observations.
However, the resulting datasets are often small in size compared to the standards adopted
by the CL community. The massive volumes of data that have become available from
sources such as social media platforms have provided the opportunity to investigate language variation more broadly.
The opportunity for the field of  sociolinguistics is to identify questions that  this massive but messy data would enable them to answer. Sociolinguists must then also select an appropriate methodology.  However, typical methods used within sociolinguistics would require sampling the data down.  If they take up the challenge to instead analyze the data in its massive form, they may find themselves open to partnerships in which they may consider approaches more typical in the field of CL.

As more and more researchers in the field of CL seek to interpret language from a social perspective, an increased awareness of insights from the field of sociolinguistics could inspire modeling refinements and potentially lead to performance gains.  
Recently, various  studies  \cite{hovyacl2015,E14-1034,volkova2013exploring} have demonstrated
that existing NLP tools can be improved by accounting for linguistic variation due to social factors, and \namecite{hovyage2015} have drawn attention to the fact that biases in
frequently used corpora, such as the Wall Street Journal, cause NLP tools to perform better on texts written by older people. The rich repertoire of theory and practice developed by sociolinguists could impact the field of CL also in more fundamental ways. The boundaries of communities are often not as clear-cut as they may seem and the impact of agency has not been sufficiently taken into account in many computational studies. For example, an understanding of linguistic agency can explain why and when there might be more or less of a problem when making inferences about people based on their linguistic choices. This issue is discussed in depth in some recent computational work related to gender, specifically 
\namecite{bamman2014gender} and \namecite{nguyen142014gender} who provide a critical reflection on the operationalization
of gender in CL studies.

The increasing interest in analyzing and modeling the social dimension of language within CL encourages collaboration between sociolinguistics and CL in various ways.
 However, the potential for synergy  between the two fields  has not been  explored systematically so far \cite{eisenstein2013bad}
and to date there is no overview  of the common  and complementary aspects of the two fields.
This article aims to present an integrated overview of research published in the two communities and to describe
 the state-of-the-art in  the emerging multidisciplinary field that could be labeled as `\emph{Computational Sociolinguistics}'. 
  The envisaged audiences are CL researchers interested in sociolinguistics and sociolinguists interested in computational
  approaches to study language use. 
We hope to demonstrate that there is enough substance to warrant the recognition of `\emph{Computational Sociolinguistics}' as an autonomous yet multidisciplinary research area. Furthermore, we hope to convey that this is the moment to develop a research agenda for the scholarly community that maintains links with both sociolinguistics and computational linguistics.

In the remaining part of this section, we discuss the rationale and scope of our survey in more detail as well as the potential impact of integrating the social dimensions of  language use in the development of practical NLP applications.
In Section~\ref{sec:methods} we discuss \emph{Methods for Computational Sociolinguistics},
in which we reflect on methods used in sociolinguistics and computational linguistics.
In Section~\ref{sec:identity}, \textit{Language and Social Identity Construction}, we discuss
how speakers use language to shape perception of their identity and focus on computational approaches to model
language variation based on  gender, age and geographical location.
In Section~\ref{sec:interaction} on \emph{Language and Social Interaction}, we move  from individual speakers to pairs, groups and communities and discuss the role of language in shaping personal relationships,
the use of style-shifting, and the adoption of norms and language change in communities.
In Section~\ref{sec:multilingualism} we discuss \emph{Multilingualism and Social Interaction},
in which we present an overview of tools for processing multilingual communication, such as parsers and language identification systems. We will also discuss approaches  for analyzing patterns in multilingual communication from a computational perspective.
In Section \ref{sec:conclusion} we conclude with a summary of major challenges within this emerging field.

\subsection{Rationale for a Survey of Computational Sociolinguistics}
The increased interest in studying a social phenomenon such as language use from a data-driven or computational perspective exemplifies a more general trend in scholarly agendas.  The study of social phenomena through computational methods is commonly referred to as  `Computational Social Science' \cite{lazer2009life}.
The increasing interest of social scientists in computational methods can be regarded as illustrating the general increase of attention for cross-disciplinary research perspectives. 
`Multidisciplinary',
`interdisciplinary', `cross-disciplinary' and `transdisciplinary' are among the labels used to mark the shift from monodisciplinary research formats to models of collaboration that embrace diversity in the selection of data and methodological frameworks.  However,  in spite of various attempts to harmonize terminology, the adoption of such labels is often poorly supported by definitions and they tend to be used interchangeably.
The objectives of research rooted in multiple disciplines often include the ambition to resolve real world or complex problems, to provide different perspectives on a problem, or to create cross-cutting research questions, to name a few \cite{Choi_etal_2006}.

The emergence of research agendas for (aspects of) computational sociolinguistics fits in this trend. We will use the term \emph{Computational Sociolinguistics} for the emerging research field that integrates aspects of sociolinguistics and computer science in studying  the relation between language and society from a computational perspective. This survey article aims to show the potential of leveraging massive amounts of data to study social dynamics in language use by combining advances in computational linguistics and machine learning with foundational concepts and insights from sociolinguistics. Our goals for establishing Computational Sociolinguistics  as an independent research area include the development of tools to support sociolinguists,  the  establishment of new statistical methods
for the modeling and analysis of data that contains linguistic content as well as information on the social context, and  the development or refinement of NLP tools based on sociolinguistic insights.

\subsection{Scope of Discussion}
Given the breadth of this field, we will limit the scope of this survey as follows.
First of all, the  coverage of sociolinguistics topics will be selective and primarily determined by the work within computational linguistics that touches on
sociolinguistic topics.
For readers with a wish for a more complete overview of sociolinguistics,
we recommend the  introductory readings by \namecite{bell2013guidebook,holmes2013introduction} and \namecite{meyerhoff}. 

 The availability of social media and other online language data in computer-mediated formats is one of the primary driving factors for the emergence of computational sociolinguistics.
A relevant research area is therefore the study of Computer-Mediated Communication (CMC) \cite{herring1996computer}.
Considering the strong focus on speech data within sociolinguistics,
there is much potential for computational approaches to be applied to  spoken language as
well. Moreover, the increased availability
of recordings of spontaneous speech and transcribed speech has inspired a revival in
the study of the social dimensions of spoken language  \cite{jain2012unsupervised}, as well as in the analysis of the relation between the verbal and the nonverbal layers in spoken dialogues \cite{Truong01122014}. As online data increasingly becomes multimodal, for example with the popularity of vlogs (video blogs), we expect the use of spoken word data for computational sociolinguistics to increase. Furthermore, we expect that multimodal analysis, a topic that has been the focus of attention in the field of human-computer interaction for many years, will also receive attention in computational sociolinguistics. 

In the study of communication in pairs and groups, the individual contributions are often 
analyzed in context.  Therefore, much of the work on language use in settings with multiple speakers draws from foundations in
discourse analysis \cite{discourseidentity2006,Hyland2004,martin2005,Schegloff}, pragmatics (such as  speech act theory  \cite{austin1975things,searle1969speech}), rhetorical structure theory \cite{mann1988rhetorical,Taboada01082006}
and social psychology \cite{giles1991language,HCRE:HCRE341,Richards2006}. For studies  within the scope of 
computational sociolinguistics that build upon these fields the link with the foundational frameworks will be indicated. 
Another relevant field is computational stylometry  \cite{Daelemans:2013fk,HOLMES01091998,ASI:ASI21001},
which focuses on computational models of writing style for various tasks such as plagiarism detection,
author profiling and authorship attribution. Here we limit our discussion to publications
on topics such as the link between style and social variables.

\subsection{NLP Applications}
Besides yielding new insights into language use in social contexts,
research in computational sociolinguistics could potentially also impact
the development of applications for the processing of textual social media and other content.
For example, user profiling tools might benefit from
research on automatically detecting the gender \cite{Burger:2011:DGT:2145432.2145568}, age \cite{ICWSM135984},
geographical location \cite{eisenstein2010latent}
or  affiliations of users \cite{E14-1012}
based on an analysis of their linguistic choices. The cases for which the interpretation of the language used could benefit most from using
variables such as age and gender are usually also the ones for which it is most difficult to automatically detect those variables.
 Nevertheless, in spite of this kind of challenge, there are some published proofs of concept that suggest potential value in advancing past the typical assumption of homogeneity of language use embodied in current NLP tools. For example, incorporating how language use varies across social groups has improved
word prediction systems \cite{E14-1034}, algorithms for cyberbullying detection \cite{dadvar2012improved}
and sentiment-analysis tools \cite{hovyacl2015,volkova2013exploring}.
\namecite{hovyage2015} show that POS taggers trained on well-known corpora such as the
English Penn Treebank perform better on texts written by older authors. They draw attention
to the fact that texts  in various frequently used corpora are from a biased
sample of authors in terms of demographic factors.
Furthermore, many NLP tools currently assume that the input consists of monolingual text,
but this assumption does not hold in all domains. For example, social media users may
employ multiple language varieties, even within a single message.
To be able to  automatically process these texts, NLP tools that are able to deal with multilingual texts are needed \cite{solorio2008part}.

\section{Methods for Computational Sociolinguistics}
\label{sec:methods}
As discussed, one important goal of this article is to stimulate collaboration between the fields of sociolinguistics in particular and social science research related to communication at large  on the one hand, and computational linguistics on the other hand.  By addressing the relationship with methods from both sociolinguistics and the social sciences in general we are able to underline two expectations.  First of all, we are convinced that sociolinguistics and related fields can help the field of computational linguistics to build richer models that are more effective for the tasks they are or could be used for.  Second, the time seems right for the CL community to contribute to sociolinguistics and the  social sciences, not only by developing and adjusting tools for sociolinguists, but also by refining the theoretical models within sociolinguistics using computational approaches and contributing to the understanding of the social dynamics in natural language.  
In this section, we highlight challenges that reflect the current state of the field of computational linguistics. 
In part these challenges relate to the fact that in the field of language technologies at large, the methodologies of social science research are usually not valued, and therefore also not taught. There is a lack of familiarity with methods that could easily be adopted if understood and accepted. 
However, 
there are promising examples of bridge building that are  already occurring in related fields such as learning analytics. More specifically, in the emerging area  of discourse analytics  there are demonstrations of how these practices could eventually be observed within the language technologies community as well \cite{discanal3,discanal2,discanal1}.

At the outset of multidisciplinary collaboration, it is necessary to understand differences in goals and values between communities, as these differences strongly influence what counts as a contribution within each field, which in turn influences what it would mean for the fields to contribute to one another.  Towards that end, we first discuss the related but distinct notions of reliability and validity, as well as the differing roles these notions have played in each field (Section \ref{sec:validation}).  This will help lay a foundation for exploring differences in values and perspectives between fields.  Here, it will be most convenient to begin with quantitative approaches in the social sciences as a frame of reference.  In Section \ref{sec:empiricism}  we discuss contrasting notions of theory and empiricism as well as the relationship between the two, as that will play an important and complementary role in addressing the concern over differing values. 
In  Section \ref{sec:qualitative_quantitative}  we broaden the scope to the spectrum of research approaches within the social sciences, including strong quantitative and strong qualitative approaches, and the relationship between CL and the social disciplines involved.  This will help to further specify the concrete challenges that must be overcome in order for a meaningful exchange between communities to take place.  In Section \ref{sec:methods_data} we illustrate how these issues come together in the role of data, as the collection, sampling, and preparation of data are of central importance to the work in both fields.
\subsection{Validation of Modeling Approaches}
\label{sec:validation}
The core of much research in the field of computational linguistics, in the past decade especially, is the development of new methods for computational modeling, such as probabilistic graphical models and deep learning within a neural network approach.  These novel methods are valued both for the \emph{creativity} that guided the specification of novel model structures and the corresponding requirement for new methods of inference as well as the achievement of \emph{predictive accuracy} on tasks for which there is some notion of a correct answer.  

Development of new modeling frameworks is part of the research production cycle both within sociolinguistics (and the social sciences in general) and the CL community, and there is a lot of overlap with respect to the types of methods used. For example, logistic regression is widely employed by variationist sociolinguists using a program called VARBRUL \cite{tagliamonte2006analysing}.  
Similarly, logistic regression is widely used in the CL community, especially in combination with 
  regularization methods when dealing with thousands of variables, for example for age prediction \cite{ICWSM135984}. As another example, latent variable modeling approaches \cite{koller2009probabilistic} have grown in prominence within the CL community for dimensionality reduction, managing heterogeneity in terms of multiple domains or multiple tasks \cite{zhang2008flexible}, and approximation of semantics \cite{Blei:2003:LDA:944919.944937,Griffiths06042004}. Similarly, it has grown in prominence within the quantitative branches of the social sciences for modeling causality \cite{glymour1987discovering}, managing heterogeneity in terms of group effects and subpopulations \cite{collins2010latent}, and time series modeling \cite{Rabe-Hesketh:2012:MLM:2361545,rabe2004gllamm}.

The differences in reasons for the application of similar techniques are indicative of differences in values.  While in CL there is a value placed on creativity and predictive accuracy, within the social sciences, the related notions of \emph{validity} and \emph{reliability} underline the values placed on conceptual contributions to the field. Validity is primarily a measure of the extent to which a research design isolates a particular issue from confounds so that questions can receive clear answers.  This typically requires creativity, and frequently research designs for isolating issues effectively are acknowledged for this creativity in much the same way a novel graphical model would be acknowledged for the elegance of its mathematical formulation.  Reliability, on the other hand, is primarily a measure of the reproducibility of a result and 
might seem to be a distinct notion from predictive accuracy. However, the connection is apparent when one considers that a common notion of reliability is the extent to which two human coders would arrive at the same judgment on a set of data points, whereas predictive accuracy is the extent to which a model would arrive at the same judgment on a set of data points as a set of judgements decided ahead of time by one or more humans. 

While at some deep level there is much in common between the goals and values of the two communities, the differences in values signified by the emphasis on creativity and predictive accuracy on the one side and reliability and validity on the other side nevertheless poses challenges for mutual exchange.
Validity is a multi-faceted notion, and it is important to properly distinguish it from the related notion of reliability.  If one considers shooting arrows at a target, one can consider reliability to be a measure of how much convergence is achieved in location of impact of multiple arrows.  On the other hand, validity is the extent to which the point of convergence centers on the target.  Reproducibility of results is highly valued in both fields, which requires reliability wherever human judgment is involved, such as in the production of a gold standard \cite{Carletta:1996:AAC:230386.230390,DiEugenio:2004:KSS:1005380.1005385}. However, before techniques from CL will be adopted by social science researchers, standards of validation from the social sciences will likely need to be addressed \cite{Krippendorff2013}. We will see that this notion requires more than the related notion of creativity as appreciated within the field of CL.

One aspect that is germane to the notion of validity that goes beyond pure creativity is the extent to which the essence that some construct actually captures corresponds to the intended quantity. This aspect of validity is referred to as \textit{face validity}.  For example, the face validity of a sentiment analysis tool could be tested as follows. First, an automatic measure of sentiment would be applied to a text corpus. Then, texts would be sorted by the resulting sentiment scores and the data points from the end points and middle compared with one another.  Are there consistent and clear distinctions in sentiment between beginning, middle, and end?  Is sentiment the main thing that is captured in the contrast, or is something different really going on? While the CL community has frequently upheld high standards of reliability, it is rare to find work that deeply questions whether the models are measuring the right thing.  Nevertheless, this deep questioning is core to high quality work in the social sciences, and without it, the work may appear weak.

Another important notion is \textit{construct validity}, or the extent to which the experimental design manages extraneous variance effectively. If the design fails to do so, it affects the interpretability of the result.  This notion applies when we interpret the learned weights of features in our models to make statements about language use. When not controlling for confounding variables, the feature weights are misleading and valid interpretation is not possible. For example, many studies on gender prediction (see Section~\ref{sec:identity}) ignore extraneous variables such as age, while gender and age are known to interact with each other highly.  Where confounds may not have been properly eliminated in an investigation, again the results may appear weak regardless of the numbers associated with the measure of predictive accuracy.

Another important methodological idea is triangulation. Simply put, it is the idea that if you look at the same object through different lenses, each of which is designed to accentuate and suppress  different kinds of details, you get more information than if you looked through just one, analogous to the value obtained through the use of ensemble methods like \textit{bagging}. 
Triangulation is thus an important way of strengthening research findings in the social sciences by leveraging multiple views simultaneously rather than just using one in addressing a question.  Sentiment analysis can again be used for illustration purposes. Consider a blog corpus for which the age of each individual blogger is available.
Let's assume that  a model  for predicting age allocated high weights  to some sentiment-related words.  This may be considered as evidence that the model is consistent with previous findings that older people use more words that express a positive sentiment. Another method could measure sentiment for each blog individually.  If the measured sentiment would correlate with the age of bloggers across the corpus,
the two methods for investigating the connection between age and sentiment would tell the same story and the confidence in the validity of the story would  increase. This type of confirming evidence is referred to as 
an indication of convergent validity.

Another form of triangulation is where distinctions known to exist are confirmed.  For this example, assume that a particular  model for predicting political affiliation placed high weights on some sentiment-related words in a corpus related to issues for which those affiliated with one political perspective would take a different stance than those affiliated with another perspective, and this affiliation is known for all data points.  The experimenters may conclude that this evidence is consistent with previous findings suggesting that voters express more positive sentiment towards political stances they are in favor of.  If this is true, then if the model is applied to a corpus where both parties agree on a stance, the measure of sentiment should become irrelevant.  Assuming the difference in the role of sentiment between the corpora is consistent with what is expected, the interpretation is strengthened.  This is referred to as divergent validity since an expected difference in relationship is confirmed.  Seeking convergent and divergent validity is a mark of high quality work in the social sciences, but it is rare in evaluations in the field of CL, and without it, again, the results may appear weak from a social science perspective.
In order for methods from CL to be acceptable for use within the social sciences, these perceived weaknesses must be addressed.

\subsection{Theory versus Empiricism}
\label{sec:empiricism}
Above we discussed the importance placed on validity within the social sciences that stems from the goal of isolating an issue in order to answer questions.  In order to clarify why that is important, it is necessary to discuss the value placed on theory versus empiricism.

Within the CL community, a paradigm shift took place after the middle of the 1990s. Initially, approaches that combined symbolic and statistical methods were of interest \cite{klavans1996balancing}. But with the focus on very large corpora and new frameworks for large-scale statistical modeling, symbolic- and knowledge-driven methods have been largely left aside, though the presence of linguistics as an active force can still be seen in some areas of computational linguistics, such as tree banking. 
 Along with older symbolic methods that required carefully crafted grammars and lexicons, the concept of knowledge source has become strongly associated with the notion of theory, which is consistent with the philosophical notion of linguistic theory advocated by Chomskyan linguistics and other formal linguistic theories
  \cite{backofen1993complete,green1992meaning,schneider2004robust,wintner2002formal}.  As knowledge-based methods have  to a large extent been replaced with statistical models, a grounding in linguistic theory has become less and less valued. A desire to replace theory with empiricism dominated the Zeitgeist and drove progress within the field.  Currently, the term \emph{theory} seems to be associated with old and outdated approaches. It often has a negative connotation in contrast to the positive reception of empiricism,
and contemporary modeling approaches are believed to have a greater ability to offer insights into language than symbolic modeling frameworks. 

In contrast, in the social sciences the value of a contribution is measured in terms of the extent to which it contributes towards theory. Theories may begin with human originated ideas.  But these notions are only treated as valuable if they are confirmed through empirical methods.  As these methods are applied, theoretical models gain empirical support.  Findings are ratified and then accumulated.  Therefore, theories  become storehouses for knowledge obtained through empirical methods.  Atheoretical empiricism is not attractive within the social sciences where the primary value is on building theory and engaging theory in the interpretation of models.  

As CL seeks to contribute to sociolinguistics and the social sciences, this divide of values must be addressed in order to avoid the fields talking at cross purposes.  To stimulate  collaboration between fields, it is important to not only focus on task performance, but also to integrate existing theories into the computational models and use these models to refine or develop new theories.

\subsection{Quantitative versus Qualitative Approaches}
\label{sec:qualitative_quantitative}
The social sciences have both strong qualitative and quantitative branches. Similarly, sociolinguistics has branches in qualitative research (e.g., interactional sociolinguistics) and quantitative research (variationist sociolinguistics). From a methodological perspective, most computational sociolinguistics work has a strong resemblance with quantitative and therefore variationist sociolinguistics, which has  a strong focus on statistical analysis to uncover the distribution of sociolinguistic variables \cite{tagliamonte2006analysing}.  So far we have mostly reflected on methods used in CL and their commonality with the methods used in the quantitative branches in sociolinguistics and the social sciences, but the time is right for a greater focus on how qualitative methods may also be of use.  Some thoughts about what that might look like can be found in the work of  \namecite{discanal2}, who explore the productive tension between the two branches as it relates to interaction analysis.  The field of computational linguistics could benefit from exploring this tension to a greater degree in its own work, for example by taking a deeper look at data through human eyes as part of the validation of constructed models.

The tension between qualitative and quantitative branches can be illustrated with the extent to which the agency of speakers is taken into account. As explained in the introduction, linguistic agency refers to the freedom of speakers to make choices about how they present themselves in interaction.  A contrasting notion is the extent to which social structures influence the linguistic choices speakers make.  Regardless of research tradition, it is acknowledged that speakers both have agency and are simultaneously influenced by social structures.  The question is which is emphasized  in the research approach.  Quantitative researchers believe that the most important variance is captured by representation of the social structure.  They recognize that this is a simplification, but the value placed on quantification for the purpose of identifying causal connections between variables makes the sacrifice of accuracy worth it.  In the field of CL, this valuing is analogous to the well-known saying that all models are wrong, but some are nevertheless useful.  On the other side are researchers committed to the idea that the most important and interesting aspects of language use are the ones that violate norms in order for the speaker to achieve a goal.  These researchers may  doubt that the bulk of choices made by speakers can be accounted for by social structures.  
We see the balance and tension between the ideas of language reflecting established social structures and language  arising from speaker agency within current trends in variationist sociolinguistics.  Much of that work focused on the ways in which language variation can be accounted for by reference to social structures \cite{bell2013guidebook}.  On the other hand, more recently the agency of speakers is playing a more central role as well in variationist sociolinguistics  \cite{eckert2012three}.

While in CL qualitative research is sometimes dismissed as being quantitative work that lacks rigor, one could argue that high quality qualitative research has a separate notion of rigor and depth that is all its own \cite{morrowbrown}.  An important role for qualitative research is to challenge the operationalizations constructed by quantitative researchers.  
To achieve the adoption of CL methods and models by social science researchers, the challenges from the qualitative branches of the social sciences will become something to consider carefully.

As computational linguistics shares more values with variationist sociolinguistics, many studies within computational sociolinguistics also focus on the influence of social structures. For example, work on  predicting social variables such as gender (Section~\ref{sec:identity}) is built on the idea that  gender determines the language use of speakers. However, such research ignores the agency of speakers:  Speakers use language to construct their identity and thus not everyone might write in a way that reflects their biological sex. Moving forward, it would make sense for researchers in computational sociolinguistics to reflect on the dominant role of social structures over agency.  Some work in CL has already begun to acknowledge the agency of speakers when interpreting findings \cite{bamman2014gender,nguyen142014gender}.

One way of conceptualizing the contrast between the usage of computational models in the two fields is to reconsider the trade-off between maximizing interpretability  --- typical of the social sciences and sociolinguistics ---,  and maximizing predictive accuracy, typical of CL. Both fields place a premium on rigor in evaluation and generalization of results across datasets. To maintain a certain standard of rigor, the CL community has produced practices for standardization of metrics, sampling, and avoidance of overfitting or overestimation of performance through careful separation of training and testing data at all stages of model development. Within the social sciences, the striving for rigor  has also produced  statistical machinery for analysis, but most of all it has resulted in an elaborate process for validation of such modeling approaches and practices for careful application and interpretation of the results.  

One consequence of the focus on interpretability within the social sciences is that models tend to be kept small and simple in terms of the number of parameters, frequently no more than 10, or at least no more than 100.  Because the models are kept simple, they can be estimated on smaller datasets, as long as sampling is done carefully and extraneous variance is controlled. In the CL community, it is more typical for models to include tens of thousands of parameters or more. For such large models, massive corpora are needed to prevent overfitting. As a result, research in the CL community is frequently driven by the availability of large corpora, which explains the large number of recent papers on data from the web, such as Twitter and Wikipedia.  Because of this difference in scale, a major focus on parallelization and approximate inference has been an important focus of work in CL \cite{heskes2002approximate}, whereas interest in such methods has only recently grown within the social sciences.

\subsection{Spotlight on Corpora and Other Data}
\label{sec:methods_data}
Data collection is a fundamental step in the research cycle for researchers in both sociolinguistics and computational linguistics. Here we will reflect on the differences in the practices and traditions within both fields and on the emerging use of online data. In the subsequent sections of this survey, there will be dedicated subsections about the data sources used in the specific studies relevant to the discussed themes (e.g., on identity construction).

Traditionally, sociolinguists have been interested in datasets that capture informal speech (also referred to as the `\textit{vernacular}'), i.e., the kind of language used when speakers are  not paying attention \cite{tagliamonte2006analysing}.
A variety of methods have been used to collect data, including observation, surveys and  interviews \cite{datacollection,tagliamonte2006analysing}. 
The sociolinguistic datasets are carefully prepared to enable in-depth analyses of how a speech community operates, carefully observing standards of reliability and validity as discussed previously.  Inevitably, these data collection methods are labor-intensive and time-consuming. The resulting datasets are often small in comparison to the ones used within computational linguistics.  The small sizes of these datasets made the work in sociolinguistics of limited interest to the field of CL.  

The tide began to turn with the rise of computer mediated communication (CMC). \namecite{herring2007} defines CMC as `\textit{predominantly text-based human-human interaction mediated by networked computers or mobile telephony}'.
The content generated in CMC, and in particular when generated on social media platforms, is a rich and easy to access source of
large amounts of informal language  coming together with  information about the context (e.g., the users, social network structure, the time or geolocation at which it was generated)
that can be used for the study of  language in social contexts on a large scale.
Examples include microblogs \cite{10.1371/journal.pone.0113114,ICWSM124661}, web forums \cite{garley2012beefmoves,Nguyen:2011:LUR:2021109.2021119} and online review sites \cite{Danescu-Niculescu-Mizil:2013:NCO:2488388.2488416,hovywww2015}.
For example, based on data from Twitter (a popular microblogging site) dialectal variation has been mapped  using a fraction of the time, costs and effort that was needed in traditional studies \cite{doyle2014mapping}. However, data from CMC is not always easy to collect. As an example, while text messaging (SMS) is widely used, collecting SMS data has been difficult due to both technical and privacy concerns. The SMS4science project \cite{sms4science} aims to overcome these difficulties by asking people to donate their messages,  collaborating with the service providers for the collection of the messages, and applying anonymization to ensure privacy.

 A complicating issue in data collection in sociolinguistics is that participants might adjust their language use towards the expectations of the data collector.
This phenomenon is known as the `observer's paradox', a term first coined by  \namecite{labov1972}:
"\textit{the aim of linguistic research in the community must be to find out how people talk when they are not being systematically
observed; yet we can only obtain these data by systematic observation}". In social media,
the observer's paradox could potentially be argued to have lost much of its strength, making it
a promising resource to complement traditional data collection methods.
While a convenient source of data, the use of social media data does introduce new challenges that must be addressed regardless of field, and this offers a convenient beginning to a potential exchange between fields.

First, social media users are usually not representative of the general population \cite{mislove2011understanding,ICWSM135984}. A better understanding of the demographics could aid the interpretation of findings, but often little is known about the users. Collecting demographic information requires significant effort, or might not even be possible in some cases due to ethical concerns.
Furthermore, in many cases the complete data is not fully accessible through an API,
requiring researchers to apply a sampling strategy (e.g., randomly, by topic, time, individuals/groups, phenomenon  \cite{onlinedatacollection,herring2004}).  Sampling may introduce additional biases or remove important contextual information. These problems are even more of a concern when datasets are reused for secondary analysis by other researchers whose purposes might be very different from those who performed the sampling.

Social media data also introduces new units of analysis (such as messages and threads) that do not correspond entirely with traditional analysis units (such as sentences and turns) \cite{onlinedatacollection}. This raises the question about valid application of findings from prior work.  Another complicating factor is that  in social media the target audience of a message is often not explicitly indicated, i.e., multiple audiences (e.g., friends, colleagues) are collapsed into a single context  \cite{marwick2011tweet}. Some studies have therefore treated the use of hashtags and user mentions as proxies for the target audience \cite{nguyenicwsm2015,pavalanathan2015linguistic}.
Furthermore, while historically the field of sociolinguistics started with a major focus on phonological variation, e.g., \namecite{labov1966}, the use of social media data  has led to a higher focus on lexical variation in computational sociolinguistics. However, there are concerns that a focus on lexical variation without regard to other aspects may threaten the validity of conclusions.
Phonology does impact social media orthography at both the word level and structural level \cite{eisensteinphonological}, suggesting that studies on phonological variation could inform studies based on  social media text data and vice versa. For example, \namecite{eisensteinphonological} found that consonant cluster reduction (e.g., \emph{just} vs. \emph{jus}) in Twitter is influenced by the phonological context, in particular, reduction was less likely when the word was followed by a segment that began with a vowel.

There are  practical concerns as well.  First, while both access and content have often been conceptualized as either public or private,
in reality this distinction is not as absolute, for example, a user might discuss a private topic on a public social media site. In view of the related privacy issues, \namecite{Bolander201414} argue for more awareness regarding the ethical implications of research using social media data.

Automatically processing social media data is more difficult compared to various other types of data that have been used within computational linguistics. Many developed tools (e.g., parsers, named entity recognizers)
do not work well due to the informal nature of many social media texts. While the dominant response has been to focus on text normalization and domain adaptation,   \namecite{eisenstein2013bad}  argues that doing so is throwing away meaningful variation.
For example, building on work on text normalization, \namecite{W11-0704} showed how various transformations
(e.g., dropping the last character of a word) vary across different user groups on Twitter.
As another example, \namecite{brody-diakopoulos:2011:EMNLP} find that lengthening of words (e.g., \emph{cooooll}) is often applied to subjective words. They build on this observation to detect sentiment-bearing words.
The tension between normalizing and preserving the variation in text also arises in the processing and analysis of historical texts (see \namecite{doi:10.2200/S00436ED1V01Y201207HLT017} for an overview), which also contain many spelling variations. In this domain, normalization is often applied as well to facilitate the use of tools such as parsers. However, some approaches first normalize the text, but then replace the modernized word forms with the original word forms to retain the original text. Another issue with social media data is that many social media studies have so far focused primarily on one data source. A comparison of the  online data sources in terms of language use has only
been done in a few studies \cite{baldwin2013noisy,hu2013dude}.

Another  up and coming promising  resource for studying language from a social perspective is crowdsourcing.
So far, crowdsourcing is mostly used to obtain large numbers of annotations, e.g., \namecite{snow2008cheap}. However, 
 `crowds' can also be used for large-scale perception studies, i.e., to study how non-linguists interpret messages and identify social characteristics of speakers \cite{datacollection-experiments}, and for the collection of linguistic data, such as the use of variants of linguistic variables. Within sociolinguistics, surveys have been one of the instruments to collect data and crowdsourcing is an emerging alternative to traditional methods for collecting survey data.

Crowdsourcing has already been used to obtain perception data for sociolinguistic research, for example, to study how English utterances are perceived differently across language communities \cite{makatchev2011perception} and
to obtain  native-likeness ratings of speech samples \cite{wieling2014}.
For some studies, games have been developed to collect data.
\namecite{nguyen142014gender} studied how  Twitter users are perceived based on their tweets by asking players to guess the gender and age based on displayed tweets. 
\namecite{10.1371/journal.pone.0143060} developed a mobile app that predicted the user's location based on a 16-question survey. By also collecting user feedback on the predictions, the authors compared their data with  the Linguistic Atlas of German-speaking Switzerland, which was collected about 70 years before the crowdsourcing study. The mismatches between the Atlas data and self-reported data from the mobile app were seen to suggest linguistic change in progress.

Crowdsourcing also introduces challenges. For example, the data collection method is less controlled and additional effort for quality control is often needed.  Even more problematic is that usually little is known is about the workers, such as the communities they are part of. For example, \namecite{wieling2014} recruited participants using e-mail, social media and blogs, which resulted in a sample that was likely to be biased towards linguistically interested people. However, they did not expect that the
possible bias in the data influenced the findings much. Another concern is that participants in crowdsourcing studies might modulate their answers towards what they think is expected, especially when there is a monetary compensation. 
In the social sciences in general, crowdsourcing is also increasingly used for survey research. \namecite{Behrend2011} compared the data collected using crowdsourcing with data collected from a traditional psychology participant pool (undergraduates)  in the context of organizational psychology research and concluded that crowdsourcing is a potentially viable resource to collect data for this research area. While thus promising, the number of studies so far using crowdsourcing for sociolinguistic research is small and more research needs to be done to study the strengths and weaknesses of this data collection method for sociolinguistic research.
\section{Language and Social Identity}
\label{sec:identity}
We now turn to discussing computational approaches for modeling language variation related to social identity.
Speakers use language to construct their social identity  \cite{bucholtz2005identity}. Being involved in  communicative exchange can be functional for the transfer of information, but at the same it functions as a staged performance in which users select specific codes (e.g., language, dialect, style)  that shape  their communication \cite{Wardhaugh2011}. Consciously or unconsciously speakers adjust their performance to the specific social context and to the impression they intend to make on their audience.
Each speaker has a personal linguistic repertoire 
to draw linguistic elements or codes from.  Selecting from the repertoire is partially subject to `identity work', a term referring to
the range of activities that individuals engage in to create, present, and sustain personal identities that are congruent with and supportive of the self-concept \cite{snow1987identity}.

Language is one of the instruments that speakers use in shaping their identities, but there are  limitations (e.g., physical or genetic constraints) to the variation that can be achieved. For example, somebody with a smoker's voice may not be able to speak with a smooth voice but many individual characteristics still leave room for variation. Although traditionally attributed an absolute status, personal features (e.g., age and gender) are increasingly considered social rather than biological variables. Within sociolinguistics, a major thrust of research is to uncover the relation between  social variables (e.g., gender, age, ethnicity, status) and  language use \cite{Eckert1997,eckert2013language,holmes2003,wagner2012age}.
The concept of sociolects, or social dialects, is similar to the concept of regional dialects. While regional dialects are language varieties based on geography, sociolects are based on  social groups, e.g., different groups according to social class (with labels such as `working class' and `middle class'), or according to gender or age. A study by \namecite{Guy201363} suggests that the cohesion between variables (e.g., nominal agreement, denasalization) to form sociolects is weaker than usually assumed. 
 The unique use of language by an individual is an idiolect, and this concept is in particular relevant for authorship attribution  (e.g., \namecite{Grieve01092007}).

Recognizing that language use can reveal social patterns,  many studies in computational linguistics
 have  focused on automatically
inferring social variables from text. This task can be seen as a form of automatic metadata detection that can provide information on author features.  The growing interest in trend analysis tools  is one of the drivers for the interest in the development  and refinement of  algorithms for this type of metadata detection. However, tasks such as
gender and age prediction do not only appeal to researchers and developers of trend mining tools. Various public demos have been able to attract the attention of the general public (e.g., TweetGenie\footnote{http://www.tweetgenie.nl} \cite{nguyen2014tweetgenie} and Gender Guesser\footnote{http://www.hackerfactor.com/GenderGuesser.php}), which can be attributed to a widespread interest in the entertaining dimension of the linguistic dimension of identity work.
The automatic prediction of individual features such as age and  gender based on only text is a nontrivial task. Studies that have compared the performance of humans with that of automatic systems for gender and age prediction based on  text alone found that automatic systems perform better than humans \cite{Burger:2011:DGT:2145432.2145568,ICWSM135984}.
A system based on aggregating guesses from a large number of people still predicted gender incorrectly
for 16\% of the Twitter users \cite{nguyen142014gender}.
While most studies use a supervised learning approach, a recent study by \namecite{Ardehaly2015} explored a lightly supervised approach using soft constraints. They combined  unlabeled geotagged Twitter data with soft constraints, like the proportion of people below or above 25 years in a county according to Census data, to train their classifiers.
 
Within computational linguistics, linguistic variation according to gender, age and geographical location
have received the most attention, compared to other variables 
such as ethnicity \cite{Ardehaly2015,pennacchiotti2011machine,rao2011hierarchical} and social class. Labels for variables like social class are more difficult to obtain and use because they are rarely made explicit in online user profiles that are publically available.
Only recently this direction has been explored, with occupation as a proxy
for variables like social class. Occupation labels for Twitter users have been extracted from their profile description  \cite{preoctiucpietro-lampos-aletras:2015:ACL-IJCNLP,10.1371/journal.pone.0138717,10.1371/journal.pone.0115545}. \namecite{10.1371/journal.pone.0138717} then mapped the derived occupations to income and 
 \namecite{10.1371/journal.pone.0115545} mapped the occupations to social class categories. However, these studies were limited to users with self-reported occupations in their profiles. 

Many studies have focused on individual social variables, but  these variables are
not independent.
For example, there are indications that linguistic features that are used more by males increase in frequency with age as well  \cite{argamon2007mining}. As another example, some studies have suggested that language variation across gender tends to be stronger
among younger people and to fade away with older ages \cite{barbieri2008patterns}. \namecite{Eckert1997} notes that the age considered appropriate for cultural events often differs for males and females (e.g., getting married), which influences the interaction between gender and age. 
The interaction between these variables is further complicated by the fact that in many uncontrolled settings
the  gender distribution may not be equal for different age ranges (as observed in blogs \cite{burger2006exploration} and Twitter \cite{ICWSM135984}).
Therefore, failing to control for gender while studying age (and vice versa)
can lead to misinterpretation of the findings.

In this section an overview will be presented of computational studies of language
variation related to social identity.
This section will first focus on the datasets that have been used to investigate social
identity and language variation in computational linguistics (Subsection \ref{sec:data_identity}).  
After surveying computational studies on language variation according to gender (Subsection \ref{sec:gender}),
age (Subsection \ref{sec:age}) and location (Subsection \ref{sec:geo}), we conclude with a discussion of how various NLP tasks, such as
sentiment detection, can be improved by accounting for language variation related to
the social identity of speakers (Subsection \ref{identity_improve_nlp}).

\subsection{Data Sources}
\label{sec:data_identity}
Early computational studies on social identity and language use
were based on formal texts, such as the British National Corpus \cite{argamon2003gender,koppel2002automatically}, or datasets collected from controlled settings,
such as recorded conversations \cite{singh2001pilot} and telephone conversations \cite{boulis2005quantitative,garera2009modeling,van2012streaming} where protocols were used to
coordinate the conversations (such as the topic).
With the advent of social media, a shift is observed towards more informal texts collected from uncontrolled settings.
Much of the initial work in this domain focused on blogs. The
Blog Authorship Corpus \cite{schler2006effects}, collected in 2004 from blogger.com, has been used in various
studies on gender and age \cite{argamon2007mining,gianfortoni2011modeling,goswami2009stylometric,nguyen2011author,sap2014}.
Others have created their own blog corpus from various sources including LiveJournal and Xanga
\cite{burger2006exploration,mukherjee2010improving,nowson2006identity,Rosenthal:2011:APB:2002472.2002569,sarawgi2011gender,yan2006gender}.

More recent studies 
are focusing on Twitter data, which contains richer interactions  in comparison to blogs.
\namecite{Burger:2011:DGT:2145432.2145568} created a large corpus by following links to blogs that contained
author information provided by the authors themselves. The dataset has been used in various subsequent studies \cite{bergsma2013using,van2012streaming,volkova2013exploring}.
Others created their own Twitter dataset \cite{eisenstein2011discovering,kokkos2014robust,liao2014lifetime,Rao:2010:CLU:1871985.1871993,al2012homophily}.
While early studies focused on English, recent studies have used Twitter data written in other languages as well, like
Dutch \cite{ICWSM135984}, Spanish and Russian  \cite{volkova2013exploring}, and Japanese, Indonesian, Turkish, and French \cite{ciot2013gender}.
Besides blogs and Twitter, other web sources have been explored, including  LinkedIn \cite{kokkos2014robust}, IMDb \cite{otterbacher2010inferring}, YouTube \cite{filippova2012user}, e-mails \cite{corney2002gender}, a Belgian social network site \cite{Peersman:2011:PAG:2065023.2065035} and Facebook \cite{rao2011hierarchical,sap2014,schwartz2013personality}.

Two aspects  can be distinguished that are often involved in the process of creating datasets
to study the relation between social variables and language use.

\paragraph*{Labeling} 
Datasets derived from uncontrolled settings such as social media often lack explicit
 information regarding the identity of users, such as their gender, age or location. 
Researchers have used different strategies to acquire adequate labels:

\begin{itemize}
\item \emph{User-provided information}. Many researchers utilize information provided by the social media users themselves,
for example based on explicit fields in user profiles  \cite{Burger:2011:DGT:2145432.2145568,schler2006effects,yan2006gender}, or by searching for specific patterns such as birthday announcements \cite{al2012homophily}.
While this information is probably highly accurate,
such information is often only available for a small set of users, e.g., for age, 0.75\% of the users in Twitter \cite{liao2014lifetime} and 55\% in blogs  \cite{burger2006exploration}.
Locations of users have been derived based on geotagged messages \cite{eisenstein2010latent} 
or locations in user profiles \cite{mubarakdarwish2014using}.

\item \emph{Manual annotation}. Another option is manual annotation based on personal information revealed in the text, profile information, and  public information on other social media sites \cite{ciot2013gender,ICWSM135984}.
In the manual annotation scenario, a random set of authors is annotated. However, the required effort is much higher resulting in smaller datasets and biases of the annotators themselves might influence the annotation process. Furthermore, for some users not enough information  may be available to even manually assign labels.

\item \emph{Exploiting names}.  Some labels can be automatically extracted based on the name of a person.
For example,  gender information for names can be derived from census information from the US Social Security Administration \cite{bamman2014gender,genderpower2014}, or from Facebook data \cite{ICWSM124644}. However, people who use names that are more common for a different gender will be incorrectly labeled in these cases. In some languages, such as Russian, the morphology of the names can also be used to predict the most likely gender labels \cite{volkova2013exploring}.
 However, people who do not provide their names, or have uncommon names, will remain unlabeled. In addition, acquiring labels this way has not been well studied yet for other languages and cultures and for other types of labels (such as geographical location or age).
\end{itemize}

\paragraph*{Sample selection}
In many cases, it is necessary to limit the study to a sample of persons.
Sometimes the selected sample is directly related to the way labels are obtained,
for example by only including people who explicitly  list their gender or age in their social media profile \cite{Burger:2011:DGT:2145432.2145568},  who have a gender-specific first name \cite{bamman2014gender}, or who have geotagged tweets \cite{eisenstein2010latent}. Restricting the sample, e.g., by only including geotagged tweets, could potentially lead to biased datasets. \namecite{pavalanathan-eisenstein:2015:EMNLP} compared 
geotagged tweets with tweets written by users with self-reported locations in their profile. They found that
geotagged tweets are more often written by women and younger people. 
Furthermore, geotagged tweets contain more geographically specific non-standard words.
Another approach is random sampling, or as random as possible due to restrictions of targeting a specific language  \cite{ICWSM135984}. However, in these cases the labels may not be readily available. This increases the annotation effort and in some cases it may not even be possible to obtain reliable labels.
Focused sampling is used as well, for example by starting with social media accounts related
 to gender-specific behavior (e.g., male/female hygiene products, sororities) \cite{Rao:2010:CLU:1871985.1871993}.
 However, such an approach has the danger of creating biased datasets, which could influence the prediction performance  \cite{cohen2013classifying}.

\subsection{Gender}
\label{sec:gender}
The study of gender and language variation has received much attention in sociolinguistics \cite{eckert2013language,holmes2003}.
Various studies have highlighted gender differences.
According to \namecite{tannen1991you}, women engage more in `rapport' talk, focusing on establishing connections,
while men engage more in `report' talk, focusing on exchanging information.
Similarly, according to \namecite{holmes1995}, in women's communication the social function of language is more salient, while in men's communication the referential function (conveying information) tends to be dominant.
\namecite{argamon2003gender} make a distinction between involvedness (more associated with women) and informational (more associated with men). 
However, with the increasing view that speakers use language to construct their identity, 
such generalizations have also been met with  criticism. Many of these studies rely on small sample sizes and ignore other variables (such as ethnicity, social class) and the many similarities between genders. Such generalizations contribute to stereotypes
and the view of gender as an inherent property. 

\subsubsection{Modeling Gender}
Within computational linguistics, researchers have focused primarily  on automatic gender classification based on text.
Gender is then treated as a binary variable based on biological characteristics, resulting
in a binary classification task. A variety of machine learning methods have been explored, including
SVMs \cite{boulis2005quantitative,ciot2013gender,corney2002gender,ICWSM124644,gianfortoni2011modeling,mukherjee2010improving,nowson2006identity,Peersman:2011:PAG:2065023.2065035,Rao:2010:CLU:1871985.1871993,al2012homophily}, logistic regression \cite{bergsma2013using,otterbacher2010inferring},
Naive Bayes \cite{goswami2009stylometric,mukherjee2010improving,yan2006gender}
and the Winnow algorithm \cite{Burger:2011:DGT:2145432.2145568,schler2006effects}.
However, treating gender as a binary variable based on biological characteristics assumes that  gender is fixed and is something people \emph{have}, instead of something people \emph{do} \cite{butler1990gender}, i.e.,
such a setup neglects the agency of speakers.
Many sociolinguists, together with scholars from the social sciences in general,
view gender as a social construct, emphasizing that gendered behavior is
a result of social conventions rather than inherent biological characteristics.


 \subsubsection{Features and Patterns}
Rather than focusing on the underlying machine learning models, most studies have focused on developing
predictive features.
Token-level and character-level unigrams and n-grams have been explored in various  studies
\cite{bamman2014gender,Burger:2011:DGT:2145432.2145568,ICWSM124644,sarawgi2011gender,yan2006gender}.
\namecite{sarawgi2011gender} found character-level language models to be more robust
than token-level language models.
Grouping words by meaningful classes could improve the interpretation and possibly the performance of  models.
Linguistic Inquiry and Word Count (LIWC, \namecite{pennebaker2001linguistic}) is
 a dictionary-based word counting program originally developed for the English language. It also has versions for other languages, such as Dutch \cite{zijlstra2005validiteit}. LIWC has been used in experiments on Twitter data \cite{ICWSM124644}
and blogs \cite{nowson2006identity,schler2006effects}. However, models  based on LIWC alone tend to perform
worse than unigram/ngram models \cite{ICWSM124644,nowson2006identity}.
By analyzing the developed features,   studies have shown  that males tend to use more numbers \cite{bamman2014gender}, technology words \cite{bamman2014gender} and URLs \cite{schler2006effects,ICWSM135984}, while females use more  terms referring to family and relationship issues \cite{boulis2005quantitative}. A discussion of the influence of genre and domain on gender differences is provided later in this section.

Various features based on grammatical structure have been explored,
including features capturing individual POS frequencies \cite{argamon2003gender,otterbacher2010inferring}
as well as POS patterns \cite{argamon2003gender,argamon2009automatically,bamman2014gender,schler2006effects}.
Males tend to use more prepositions  \cite{argamon2007mining,argamon2009automatically,otterbacher2010inferring,schler2006effects} and
more articles  \cite{argamon2007mining,nowson2006identity,otterbacher2010inferring,schler2006effects,schwartz2013personality}, however \namecite{bamman2014gender} did not find these differences to be significant in their Twitter study.
Females tend to use more pronouns \cite{argamon2003gender,argamon2007mining,argamon2009automatically,bamman2014gender,otterbacher2010inferring,schler2006effects,schwartz2013personality}, in particular
first person singular \cite{ICWSM135984,otterbacher2010inferring,schwartz2013personality}.
A measure introduced by \namecite{heylighen2002variation} to measure formality  based on the frequencies
of different word classes has been used in experiments on blogs  \cite{mukherjee2010improving,nowson2005weblogs}.
\namecite{sarawgi2011gender} experimented with probabilistic context-free grammars (PCFGs) by adopting the approach proposed by \namecite{raghavan-kovashka-mooney:2010:Short} for authorship attribution. 
They trained PCFG parsers for each gender and computed the likelihood of test documents for each gender-specific PCFG parser to make the prediction.
\namecite{bergsma-post-yarowsky:2012:NAACL-HLT} experimented with three types of syntax features and found features based on single-level context-free-grammar (CFG) rules (e.g.,  \texttt{NP}$\rightarrow$ \texttt{PRP}) to be the most effective.
In some languages such as French,
the gender of nouns (including the speaker) is often marked in the syntax. 
For example, a male would write `\emph{je suis all{\'e}}', while a female would  write \emph{`je suis all{\'e}e'}
(`I went').
By detecting such \emph{`je suis'} constructions, \namecite{ciot2013gender} improved performance
of gender classification in French.

Stylistic features have been widely explored as well.
Studies have reported that males tend to use longer words, sentences and texts \cite{goswami2009stylometric,otterbacher2010inferring,singh2001pilot},
and more swear words \cite{boulis2005quantitative,schwartz2013personality}.
Females use more emotion words  \cite{bamman2014gender,nowson2006identity,schwartz2013personality}, emoticons \cite{bamman2014gender,gianfortoni2011modeling,Rao:2010:CLU:1871985.1871993,volkova2013exploring},
and typical social media words such as \emph{omg} and \emph{lol}  \cite{bamman2014gender,schler2006effects}.

Groups can be characterized by their attributes, for example females tend to have maiden names.
\namecite{bergsma2013using}  used such distinguishing attributes,
extracted from common nouns for males and females (e.g., granny, waitress), to improve classification performance.
Features based on first names have also been explored. Although not revealing much about language use itself, they can improve prediction performance  \cite{bergsma2013using,Burger:2011:DGT:2145432.2145568,rao2011hierarchical}.

\paragraph*{Genre}
So far, not many studies have analyzed the influence of genre 
and domain \cite{genre2001} on language use, but
a better understanding 
will aid the interpretation of observed language variation patterns.
Using data from the British National Corpus, 
\namecite{argamon2003gender} found 
a strong correlation between characteristics of male and non-fiction writing and likewise, between female and fiction writing.
Based on this observation, they trained  separate  prediction models for fiction and non-fiction \cite{koppel2002automatically}.
Building on these findings,  \namecite{herring2006gender} investigated whether gender differences would still be observed
when controlling for genre in blogs. They did not find a significant relation between gender and linguistic features that were identified to be associated with gender in previous literature,
however the study was based on a relatively small sample.
Similarly, \namecite{gianfortoni2011modeling} revisited the task of gender prediction on the Blog Authorship Corpus. After controlling for occupation, features that previously were found to be predictive for gender on that corpus were not effective anymore.

Studies focusing on gender prediction have tested the generalizability of gender prediction models 
 by training and testing on different datasets.
Although models tend to perform worse when tested on a different dataset than the one used for training, 
studies have shown that prediction performance is still higher than random, suggesting  that there are indeed gender-specific patterns 
of language variation
that go beyond genre and domain \cite{sap2014,sarawgi2011gender}.
\namecite{gianfortoni2011modeling} proposed the use of `stretchy patterns',
flexible sequences of categories, to model stylistic variation and to improve generalizability across domains.

\paragraph*{Social Interaction}
Most computational studies on gender-specific patterns in language use have studied speakers in isolation. As the conversational partner\footnote{An individual who participates in a conversation, sometimes also referred to as
interlocutor or addressee} and social network  influence the language use of speakers, several studies have extended their focus
by also considering contextual factors. For example, this led to the finding that speakers use more gender-specific language in same-gender conversations \cite{boulis2005quantitative}.
On the Fisher and Switchboard corpus (telephone conversations), classifiers dependent on the gender of the conversation partner improve performance \cite{garera2009modeling}. However, exploiting the social network of speakers on Twitter has been less effective so far. Features derived from the friends of Twitter users did not improve gender classification (but it was effective for age) \cite{al2012homophily}.
Likewise,  \namecite{bamman2014gender} found that social network information of Twitter users did not improve gender classification when enough text was available.

Not all computational studies on gender in interaction contexts have focused on gender classification itself.
Some have used gender as a variable when studying other phenomena.
In a study on language and power, \namecite{genderpower2014}
showed how the gender composition of a group influenced how power is manifested in the Enron corpus, a large collection of emails from Enron employees (described in more detail in Section \ref{interaction_data}).
In a study on language change in online communities,
\namecite{hemphill2012learning} found that females write more like men over time in the IMDb community (a movie review site), which they explain by men receiving more prestige in the community.
\namecite{jurafsky-ranganath-mcfarland:2009:NAACLHLT09} 
automatically classified speakers according to interactional style (awkward, friendly, or flirtatious)
using various types of features, including  lexical features based on LIWC \cite{pennebaker2001linguistic},
prosodic, and discourse features.
Differences, as well as commonalities, were observed between genders, and incorporating features from both speakers improved classification performance.

\subsubsection{Interpretation of Findings}
As mentioned before, most computational approaches
adopt a simplistic view of gender as an inherent property based on biological characteristics.
Only recently, the computational linguistics community has noticed the limitations of this simplistic view
by acknowledging the agency of speakers.
Two of these studies based their argumentation on an analysis of the social networks of the users.
Automatic gender predictions on YouTube data correlated more
strongly with the dominant gender in a user's network
than the user-reported gender  \cite{filippova2012user}.
Likewise, in experiments by \namecite{bamman2014gender},  incorrectly labeled  Twitter users also had
 fewer same-gender connections. In addition, they identified clusters of users who used linguistic markers that conflicted with general population-level findings.
Another study was based on data collected from an online game \cite{nguyen142014gender}.
Thousands of players guessed the age and gender of Twitter users based on their tweets,
and the results revealed that many Twitter users
 do not tweet in a gender-stereotypical way.

Thus, language is inherently social and 
while certain language features are \emph{on average} used more by males or females,
individual speakers may diverge from the stereotypical images that tend to be highlighted by many studies.
In addition, gender is shaped differently depending on the culture and language, and thus presenting
gender as a universal social variable can be misleading.
Furthermore, linguistic variation within speakers of the same gender holds true as well.

\subsection{Age}
\label{sec:age}
Aging is a universal phenomenon and understanding the relation between language and age
can provide interesting insights in many ways. An individual at a specific
time represents both a place in history as well as a life stage \cite{Eckert1997},
and thus observed patterns can generate new insights into language change 
as well as how individuals change their language use as they move through life.
Within computational linguistics, fewer studies have focused on language variation according to age compared to studies focusing on gender,
possibly because obtaining age labels requires more effort than gender labels  (e.g., the gender of people can often be derived from their names; cf. Section~\ref{sec:data_identity}).
Most of these studies have focused on absolute chronological age, although 
age can also be seen as a social variable like gender.

Sociolinguistic studies have found that adolescents  use the most non-standard forms, because  at a young age the
group pressure to not conform to established societal conventions is the largest \cite{Eckert1997,holmes2013introduction}.
In contrast, adults are found to use the most standard language, because for them social advancement matters
and they use standard language to be taken seriously \cite{bell2013guidebook,Eckert1997}.
These insights can explain why predicting the ages of older people is harder, e.g., distinguishing between a 15- and a 20-year old person based on their language use is easier than distinguishing between  a 40- and a 45-year old person \cite{ICWSM135984,nguyen142014gender}.
Thus, age is an important variable to consider, especially when we consider processes relevant for language evolution, since the degree of language innovation varies by age \cite{labovsocial}.

\subsubsection{Modeling Age} 
A fundamental question is \emph{how} to model age, and so far researchers have not reached a consensus yet.
\namecite{Eckert1997} distinguishes between chronological age (number of years since birth), biological age (physical maturity)
and social age (based on life events). Speakers are often grouped according to their age, because the amount of data is in many cases not sufficient to make more fine-grained distinctions \cite{Eckert1997}.
Most studies consider chronological age and group speakers based on age spans \cite{barbieri2008patterns,labov1966,Trudgill1974}.
However, chronological age can be misleading since persons with the same chronological age may have had very different life experiences. Another approach is to group speakers according to `shared experiences of time', such as high school students \cite{Eckert1997}.

Within computational linguistics the most common approach is to model age-specific language use based on the chronological age of speakers. An exception is  \namecite{ICWSM135984}  who  explored classification into life stages.
However, even when focusing on chronological age, the task can be framed in different ways as well.
Chronological age prediction has mostly been approached as a \emph{classification} problem,
by  modeling the chronological age
as a \emph{categorical} variable. Based on this task formulation, various classical machine learning models have been used, such as SVMs \cite{Peersman:2011:PAG:2065023.2065035,Rao:2010:CLU:1871985.1871993}, logistic regression
\cite{ICWSM135984,Rosenthal:2011:APB:2002472.2002569} and Naive Bayes \cite{tam2009age}.

The boundaries used for discretizing age have varied depending on the dataset and experimental setup.
Experiments on the blog authorship corpus \cite{schler2006effects} used categories based on the following age spans: 13-17,
23-27, and 33-47, removing the age ranges in between to simplify the task. \namecite{rangel2013overview} adopted this approach in the Author Profiling task at PAN 2013.
The following year, the difficulty of the task at PAN 2014 was increased by considering the more fine-grained categories of
18-24, 25-34, 35-49, 50-64 and 65+ \cite{rangeloverview}.
\namecite{al2012homophily} classified Twitter users into 18-23 and 25-30.
Other studies explored boundaries at 30 \cite{Rao:2010:CLU:1871985.1871993}, at 20 and 40 \cite{ICWSM135984},
at 40 \cite{garera2009modeling} and at 18 \cite{burger2006exploration}.

In several studies experiments have been done by varying the classification boundaries. 
\namecite{Peersman:2011:PAG:2065023.2065035} experimented with binary classification and boundaries at 16, 18 and 25.  \namecite{tam2009age}  experimented with classifying
teens versus 20s, 30s, 40s, 50s and adults. Not surprisingly, in both studies a higher performance was obtained when using larger age gaps (e.g., teens versus 40s/50s) 
than when using smaller age gaps (e.g., teens versus 20s/30s) \cite{Peersman:2011:PAG:2065023.2065035,tam2009age}.
\namecite{Rosenthal:2011:APB:2002472.2002569} explored a range of splits to study differences in performance when predicting the birth year of blog authors. They related their findings to pre- and post social media generations.

 For many applications, modeling age as a categorical variable might be sufficient. However, it does have several limitations. First, selecting age boundaries has proven to be  difficult. It is not always clear which categories are meaningful. 
 Secondly, researchers have used different categories depending on the age distribution of their dataset, which makes it difficult to make comparisons across datasets.


Motivated by such limitations, recent studies have modeled age as a \emph{continuous} variable,
removing the need to define age categories.
Framing age prediction as a regression task, a frequently used method has been linear regression  \cite{nguyen2011author,ICWSM135984,sap2014,schwartz2013personality}.
\namecite{liao2014lifetime} experimented with a latent variable model that jointly models age and topics.
In their model, age-specific topics obtain low standard deviations of age, while more general topics obtain high standard deviations. Another approach that would remove the need to define age categories is the unsupervised induction of age categories. Analyzing the discovered age groups could  shed more light on the relation between language use and age, but we are not aware of existing research in this area.

 \subsubsection{Features and Patterns}
The majority of studies on age prediction have focused on identifying predictive features.
While some features tend to be effective across domains, others are domain-specific \cite{nguyen2011author}.
 Features that characterize male speech have been found to also increase with age \cite{argamon2007mining}, thus simply said, males tend to sound older than they are.

Unigrams alone already perform well \cite{nguyen2011author,ICWSM135984,Peersman:2011:PAG:2065023.2065035}. 
Features based on part of speech are effective as well.
For example, younger people tend to use more first and second person singular pronouns (e.g., \emph{I, you}),
while older people more often use first person plural pronouns (e.g., \emph{we}) \cite{barbieri2008patterns,ICWSM135984,Rosenthal:2011:APB:2002472.2002569}.
Older people also use more prepositions \cite{argamon2009automatically,ICWSM135984},
determiners \cite{argamon2009automatically,nguyen2011author}
and articles \cite{schwartz2013personality}. Most of these studies focused on English
and therefore some of these findings might not be applicable to other languages.
For example, the effectiveness of pronoun-related features should also be studied in pro-drop languages (e.g., Turkish and Spanish).

Various  studies have found that younger people use less standard language.
They use more alphabetical lengthening (e.g., \emph{niiiice}) \cite{ICWSM135984,Rao:2010:CLU:1871985.1871993},
more contractions without apostrophes (e.g., \emph{dont}) \cite{argamon2009automatically},
more Internet acronyms (e.g., \emph{lol}) \cite{Rosenthal:2011:APB:2002472.2002569},
more slang \cite{barbieri2008patterns,Rosenthal:2011:APB:2002472.2002569},
more swear words \cite{barbieri2008patterns,nguyen2011author},
and more capitalized words (e.g., \emph{HAHA}) \cite{ICWSM135984,Rosenthal:2011:APB:2002472.2002569}.
Specific words such as \emph{like} are also highly associated with younger ages \cite{barbieri2008patterns,nguyen2011author}.
Younger people also use more features that indicate stance and emotional involvement \cite{barbieri2008patterns},
such as  intensifiers \cite{barbieri2008patterns,ICWSM135984} and  emoticons \cite{Rosenthal:2011:APB:2002472.2002569}.
Younger people also use shorter words and sentences and write shorter tweets  \cite{burger2006exploration,ICWSM135984,Rosenthal:2011:APB:2002472.2002569}.

\subsubsection{Interpretation of Findings}
Age prediction experiments are usually done on datasets collected at a specific point in time.
Based on such datasets, language use is modeled and compared between users with different ages.
Features that are found to be predictive or that correlate highly with age are used to highlight how
differently `younger' and `older' people talk or write.
However, the observed differences in language use based on such datasets
could be explained in multiple ways.
Linguistic variation can occur as an individual moves through life (\emph{age grading}).
In that case the same trend is observed for individuals at different time periods.
Linguistic variation can also be a result of changes in the community itself as it 
moves through time (\textit{generational change}) \cite{bell2013guidebook,sankoff2006age}. 
For example, suppose we observe that younger Twitter users include more smileys in their tweets.
This could indicate that smiley usage is higher at younger ages, but that when Twitter users
grow older they decrease their usage of smileys. However, this could also indicate
a difference in smiley usage between generations (i.e., the generation of the current younger Twitter users
use more smileys compared to the generation of the older Twitter users).
This also points to the relation between synchronic variation
and diachronic change.
Synchronic variation is variation across different speakers or speech communities at a particular point in time,
while diachronic change is accumulation of synchronic variation in time and frequency. To have a better understanding of change, we need to understand the spread of variation across time and frequency.
As is the case for gender, age can be considered a social variable and thus when only modeling chronological age,
we are ignoring  the agency of speakers and that speakers  follow different trajectories in their lives.

\subsection{Location}
\label{sec:geo}
Regional variation has been extensively studied in sociolinguistics
and related areas such as  dialectology  \cite{chambers1998dialectology} and dialectometry \cite{doi:10.1146/annurev-linguist-030514-124930}. 
The use of certain words, grammatical constructions, or the pronunciation of a word, can often reveal where a speaker is from.
For example, `\textit{yinz}' (a form of the second-person pronoun) is mostly used around Pittsburgh, which can be observed on Twitter as well \cite{eisenstein2015geographical}.
Dialectology traditionally focuses on the geographical distribution of individual or small sets of linguistic variables  \cite{chambers1998dialectology}.
A typical approach involves identifying and plotting \textit{isoglosses}, lines that divide maps into regions where specific  values of the variable predominate. The next step involves identifying bundles of isoglosses, often followed by the identification of dialect regions. 
While these steps have usually been done manually, computational approaches have recently been explored as well. For example, \namecite{LVC:8347723} demonstrated how methods from spatial analysis can be used for automating such an analysis.

The study of regional variation has been heavily influenced by new statistical approaches, such as from computational linguistics, machine learning and spatial analysis.  A separate branch has also emerged, referred to as dialectometry \cite{doi:10.1146/annurev-linguist-030514-124930}. In contrast to dialectology, which focuses on individual linguistic variables, dialectometry involves aggregating linguistic variables to examine linguistic differences between regions. \namecite{LNC3:LNC3114} argues that studies that focus on individual variables are sensitive to noise and that therefore aggregating linguistic variables will result in more reliable signals. This aggregation step has led to the introduction of   various statistical methods, including clustering, dimensionality reduction techniques
and regression approaches  \cite{dialectometry2013,nerbonne2015,W10-2305}. Recently, researchers within dialectometry
have explored the automatic identification of characteristic features of dialect regions \cite{W10-2305}, a task which  aligns more closely with the 
approaches taken by dialectologists.

While the datasets typically used in dialectology and dialectometry studies are still small compared
to datasets used in computational linguistics, similar  statistical methods have been explored.
This has created a promising starting point for closer collaboration with computational linguistics.

\subsubsection{Modeling Geographical Variation}
\label{identity_regional}
Within CL, we find two  lines of work on computationally modeling  geographical variation.

\paragraph*{Supervised} 
The first approach starts with documents labeled according to their dialect, which
can be seen as a supervised learning approach. Most studies taking this approach focus on
automatic dialect identification, which is a variation of automatic language identification, a well-studied research topic
within the field of computational linguistics \cite{baldwin2010language,hughes2006reconsidering}. While some have considered automatic language identification a solved problem \cite{mcnamee2005language}, still many outstanding issues exist
\cite{hughes2006reconsidering}, including the identification of dialects and closely related languages \cite{zampieri2014report,zampieri-EtAl:2015:LT4VarDial1}.
In studies on automatic dialect identification, various dialects have been explored, including Arabic \cite{darwish2014verifiably,elfardy2013sentence,huang:2015:EMNLP,zaidan2013arabic}, Turkish \cite{dogruoznakovredicting},  Swiss German \cite{D10-1112} and Dutch \cite{trieschnigg2012exploration} dialects.

\paragraph*{Unsupervised}
An alternative approach  is to start with location-tagged data to automatically identify dialect regions.
While the models are given labels indicating the locations of speakers, the dialect labels themselves
are not observed. In the context of modeling dialects, we consider it an unsupervised approach (although
it can be considered a supervised approach when the task is framed as a location prediction task).
The majority of the work in this area has used Twitter data, because it contains fine-grained location
information in the form of GPS data for tweets or user-provided locations in user profiles.

Much of the research that starts with location-tagged data is done with the aim of automatically
predicting the location of speakers.
The setup  is thus similar to the setup for the other tasks that we have surveyed  in this section (e.g., gender
and age prediction).
\namecite{eisenstein2010latent} developed a topic model to identify geographically coherent
linguistic regions and words that are highly associated with these regions.
The model was tested by predicting the locations of Twitter users based on their tweets.
While the topic of text-based location prediction has received increasing attention \cite{bo2012geolocation,wing2011simple},
 using these models for the discovery of  new  sociolinguistic patterns is an option that has not been fully explored yet,
 since most studies primarily focus on prediction performance.

Various approaches have been explored to model the location of speakers,
an aspect that is essential in many of the studies that start with location-tagged data.
In \namecite{wing2011simple},  locations are modeled using geodesic grids,
but these grids do not always correspond  to administrative  or language boundaries.
Users can also be grouped based on cities \cite{bo2012geolocation}, but such an approach
is not suitable for users in rural areas or when the focus is on more fine-grained geographical variation (e.g., within a city).
\namecite{eisenstein2010latent} model regions using Gaussian distributions,
but only focus on the United States and thus more research is needed to investigate the suitability
of this approach when considering other countries or larger regions.

 \subsubsection{Features and Patterns}
Word and character n-gram models have been frequently used  in  dialect identification  \cite{king2014experiments,trieschnigg2012exploration,zaidan2013arabic}.
Similarly, many text-based location prediction systems make use of unigram word features \cite{eisenstein2010latent,bo2012geolocation,wing2011simple}.

Features inspired by sociolinguistics could potentially improve performance.
\namecite{darwish2014verifiably} showed that for identifying  Arabic dialects 
a better  classification  performance could be obtained by  incorporating  known lexical, morphological and phonological differences in their model. \namecite{D10-1112} also found that using linguistic knowledge
improves over an n-gram approach. 
Their method is based on   a linguistic atlas for the extraction of lexical, morphological and phonetic rules and the likelihood of these forms across
German-speaking Switzerland. \namecite{dogruoznakovredicting} explored the use of light verb constructions to
distinguish between two Turkish dialects. 

To support the discovery of new sociolinguistic patterns and to
improve prediction performance, several studies have focused on 
automatically identifying characteristic features of dialects.
\namecite{bo2012geolocation} explored various feature selection methods to improve location prediction. 
The selected features may reflect dialectal variation but this was not the focus of the study.
The method by \namecite{W12-0211} was based on in-group and out-group comparisons using data in which  linguistic varieties
were already grouped (e.g., based on clustering).
\namecite{peirsman2010automatic} compared
frequency-based measures, such as chi-square and log-likelihood tests, with distributional methods.
Automatic methods may identify many features that vary geographically such as
topic words and named entities, and an open challenge is to separate  this type of variation from the more sociolinguistically interesting variations. For example, the observation that the word `\textit{beach}' is used more often near coastal areas
or that `\textit{Times Square}' is used more often in New York is not interesting from the perspective of a sociolinguist.

Making use of location-tagged data, several studies have focused on analyzing patterns of regional variation.
\namecite{doyle2014mapping} analyzed the geographical distribution of dialectal variants (e.g., the use of double modals like `\emph{might could}') based on Twitter data, and compared it with traditional sociolinguistic
data collection methods. Starting with a query-based approach,
he uses baseline queries (e.g., `\textit{I}')  for estimating
 a conditional distribution of data given metadata. His approach achieved high correlations with  data from
sociolinguistic studies.
\namecite{anna2015} studied the use of three phonological features of African American Vernacular English using manually selected word pairs. The occurrence  of the features was correlated with location data (longitude and latitude) as well as demographic information obtained from the US census bureau.
While these approaches start with attested  dialect variants,
automatic discovery of unknown variation patterns could potentially lead to even more  interesting results.
 To study how a word's meaning varies geographically, \namecite{P14-2134} extended the skip gram model by \namecite{mikolov2013efficient} by adding contextual variables that represent states from the US. The model then learns a global embedding matrix and additional matrices
for each context (e.g., state) to capture the variation of a word's meaning.

The increasing availability of longitudinal data has made it possible to study the spreading of linguistic innovations
geographically and over time on a large scale.
A study by  \namecite{10.1371/journal.pone.0113114} based on tweets in the United States indicates that linguistic innovations spread through demographically similar areas, in particular with regard to race.

\subsubsection{Interpretation of Findings}
Labeling texts by dialect presumes that there are  clear boundaries between dialects. However, it is not easy to make absolute distinctions between language varieties (e.g., languages, dialects). 
 \namecite{chambers1998dialectology} illustrate this with the example of traveling from village to village in a rural area. 
Speakers from villages  at larger distances  have more difficulty understanding each other compared to villages that are closer to each other, but there is no clear-cut distance at which speakers are no longer mutually intelligible. A computational approach was taken by \namecite{heeringa2001dialect} to shed more light on this puzzling example. 
 Besides linguistic differences, boundaries between language varieties are often influenced by other factors such as political boundaries \cite{chambers1998dialectology}.  
Therefore, deciding on the appropriate labels to describe linguistic communication across different groups of speakers (in terms of language, dialect, minority language, regional variety, etc.)  is an on-going issue of debate. 
 The arbitrariness of the distinction between
 a language and dialect is captured with the popular expression  "\textit{A language is a dialect with an army and navy}" \cite{LSY:4180952}. 
Methods that do not presume clear dialect boundaries are therefore a promising alternative. However, such methods
then rely on location-tagged data, which is usually only available for a portion of the data. 

\subsection{Text Classification Informed by Identity Information}
\label{identity_improve_nlp}
So far, we have focused on automatically predicting the variables themselves (e.g., gender, age, location) but
linguistic variation related to the identity of speakers can also be used to improve various other NLP tasks.
\namecite{dadvar2012improved} trained gender-specific classifiers to detect
instances of cyberbullying, noticing that language used by harassers varies by gender.
To improve the prediction performance of detecting  
the power direction between participants in emails, \namecite{genderpower2014} incorporated the gender of participants in e-mail conversations
and the overall `gender environment' as features in their model.
\namecite{volkova2013exploring}  studied gender differences in the use of subjective language on Twitter.
Representing gender as a binary feature was not effective,
but the use of features based on gender-dependent sentiment terms improved subjectivity and polarity classification.
\namecite{hovyacl2015} found that training gender- or age-specific word embeddings improved
tasks such as sentiment analysis and topic classification.

\section{Language and Social Interaction}
\label{sec:interaction}
The previous section explored computational approaches to the study of identity construction through language. We discussed variables such as gender, age and geographical location,
thereby mostly focusing on the influence of social structures on language use. However, as we also pointed out in the previous section, speaker agency enables violations of conventional language patterns. 
Speakers do not act in isolation, but they are part of pairs, groups and communities.  Social interaction contexts  produce the opportunity for variation due to agency.
In response to the particulars of these social settings and encounters (e.g., the addressee or audience, topic, and social goals of the speakers),  there is thus much variation within individual speakers.
The variation that is related to the context of interaction will be the focus of this section.

We start this section with a discussion of data sources for large-scale analyses of language use in pairs, groups and communities (Section \ref{interaction_data}).
Next, we discuss computational approaches to studying how language reflects and shapes footing within social relationships (Section \ref{personal_relationships}). Much of this work has revolved around the role of language in power dynamics by studying how speakers use language to maintain and change power relations \cite{fairclough1989language}.
We will continue with a discussion on style-shifting (i.e., the use of different styles by a single speaker) in Section \ref{style_shifting}. We will discuss  two prominent frameworks within
sociolinguistics, Audience Design \cite{LSY:2990984} and Communication Accommodation Theory \cite{giles19911}, and discuss how these frameworks have been studied within the computational linguistics community.
Finally, we will move our focus to the community level and discuss computational studies on how members adapt their language to conform to or sometimes diverge from community norms.  One might speculate about how these micro-level processes might eventually become conventional, and therefore consider how these processes may lead to language change over time (Section \ref{community_dynamics}).

\subsection{Data Sources}
\label{interaction_data}
Many of the types of data that are relevant for the investigation of concepts of social identity, are also relevant for work on communication dynamics in pairs, groups and communities.
The availability of detailed interaction recordings in online data has driven and enabled much of the work on this topic
within computational linguistics. 
A  variety of online discussion forums have been analyzed, including online cancer support
communities \cite{Nguyen:2011:LUR:2021109.2021119,wang2014}, a street gang forum \cite{E14-1012}, and more recently discussion forums in Massive Open Online Courses (MOOCs) \cite{mooc1,mooc2}.
Review sites, such as TripAdvisor \cite{ICWSM148046}, IMDb \cite{hemphill2012learning} and beer review communities \cite{Danescu-Niculescu-Mizil:2013:NCO:2488388.2488416}, have also been used in studies on language in online communities.

The Enron email corpus is another frequently used data source. The Enron corpus is a large email corpus with messages from Enron employees, which was made public
during the legal investigation of the Enron corporation. 
The corpus has been used in various studies, for example, investigations related to email classification  \cite{Klimt2004} and structure of communication networks \cite{diesner2005}.  In particular, in studies
on language and social dynamics, the Enron email corpus has featured in analyses of power relationships \cite{Diehl:2007:RIS:1619645.1619733,Gilbert:2012:PSW:2145204.2145359,C12-1138,genderpower2014}, since
Enron's organizational structure is known and can be integrated in studies on hierarchical power structures connected with quantitative capacity theories of power. Such theories treat power as a stable characteristic that inheres in a person. An example theory within this space is Resource Dependency Theory \cite{rdp1978}.

For studies that involve more dynamic notions of power (e.g., identifying individuals who are pursuing power), other resources
have also been explored, including Wikipedia Talk Pages \cite{W11-0707,6337148,danescu2012echoes,swayamdipta2012pursuit}, 
transcripts of political debates \cite{I13-1042,D14-1157} and transcripts of Supreme Court arguments \cite{danescu2012echoes}. 

\subsection{Shaping Social Relationships}
\label{personal_relationships}
Language is not only a means to exchange information but language also contributes to the performance of action within interaction.
Language  serves simultaneously as a reflection of the relative positioning of speakers to their conversation partners as well as actions that accompany those positions \cite{CBO9780511584459A012}.   Sometimes distributions of these actions can be considered to cohere to such a degree that they can be thought of as defining conversational roles \cite{conv-roles}.
At a conceptual level, this work draws heavily from a foundation in linguistic pragmatics  \cite{grice1975,Levinson1983}
as well as sociological theories of discourse  \cite{gee2011,tannen1993}, which each provide a complementary view.
Concepts related to expectations or norms that provide the foundation for claiming such positions may similarly be described either from a philosophical perspective or a sociological one \cite{HCRE:HCRE341}. In viewing interaction as providing  a context in which information and action may flow towards the accomplishment of social goals, speakers position themselves and others as sources or recipients of such information and action \cite{martin2003}.  When performatives, i.e., speech acts used to perform an action, break norms related to social positions, they have implications for relational constructs such as politeness \cite{brown1987politeness}, which codifies rhetorical strategies for acknowledging and managing relational expectations while seeking to accomplish extra-relational goals. 
In the remaining part of this section, we focus on computational studies within this theme.
We first discuss the general topic of automatic extraction of social relationships from text, and then
focus on power and politeness.

\paragraph*{Automatic Extraction of Social Relationships}
Recognizing that language use may reveal cues about social relationships, studies within CL have explored the automatic extraction of
different types of social relationships based on text.
One distinction that has been made is between weak ties (e.g., acquaintances) and strong ties
(e.g., family and close friends)  \cite{tiestrength}. \namecite{Gilbert:2009:PTS:1518701.1518736} 
explored how different types of information (including messages posted) can be used to predict
tie strength on Facebook. In this study, the predictions were done for ties within a selected sample.
\namecite{bak-kim-oh:2012:ACL2012short} studied differences in self-disclosure
on Twitter between strong and weak ties using
automatically identified topics.
Twitter users disclose more
personal information to strong ties, but they
show more positive sentiment towards weak ties, which may be explained by social norms regarding first-time acquaintances on Twitter.

Other studies have automatically extracted social relationships
from more extensive datasets, enabling analyses of the extracted network structures.
These studies have focused on extracting
signed social networks, i.e., networks with positive and negative edges, for example based on positive and negative affinity between individuals
or formal and informal relationships. 
Work within this area has drawn from   Structural Balance Theory \cite{balance}, which captures intuitions
such as that when two individuals have a mutual friend, they are likely to be friends as well,
and from Status Theory \cite{Leskovec:2010:SNS:1753326.1753532},
which involves edges that are directed and reflect status differences.
\namecite{D12-1006} developed a machine learning classifier to extract signed social networks
and found that the extracted network structure mostly agreed with Structural Balance Theory. 
\namecite{lebowsky2015}  proposed an unsupervised model for extracting signed social networks,
which they used to extract formal and informal relations in a movie-script corpus. 
Furthermore, their model also induced the social function of address terms (e.g., \emph{dude}).
To infer edge signs in a social network, \namecite{Q14-1024} formulated an optimization problem that combined two objectives, capturing the extent to which the inferred signs  agreed with the predictions of a sentiment analysis model, and the extent to which the  resulting triangles corresponded with  Status and Structural Balance Theory.

\paragraph*{Power} Work on power relations  draws from social psychological concepts of relative power in social situations \cite{socialpsypower2010}, in particular aspects of relative power that operate at the level of individuals in relation to specific others within groups or communities.
Relative power may be thought of as operating in terms of horizontal positioning or vertical positioning:
Horizontal positioning relates to closeness and related constructs such as positive regard, trust and commitment, while vertical positioning relates to authority and related constructs such as approval and respect among individuals within communities. 
Within the areas of linguistics and computational linguistics, investigations have focused on how speakers use language to maintain and change power relations \cite{fairclough1989language}.  Operationalization  and computational modeling of these two dimensions has important applications in the field of learning sciences \cite{souffle}.

Within computational linguistics, much of the work related to analysis of power as it is reflected through language  has focused on automatically identifying power relationships from text.  Though some of the literature cited above is referenced in this work, the engagement between communities has remained so far at a simple level.  Fine-grained distinctions between families of theories of power, and subtleties about the relationship between power and language are frequently glossed over.  One way in which this is visible is in the extent to which the locus of meaning is treated as though it is in the text itself rather than an emergent property of the interaction between speakers.  Though some references to external power structures and transient power relationships are mentioned, much room remains for deeper reflection on the connection between power and language. 

Research in the computational linguistics community related to these issues is normally centered around classification tasks. 
Earlier studies have focused on hierarchical power relations based on the organizational structure,
thereby frequently making use of the Enron corpus.
 \namecite{bramsen2011extracting} extracted messages
between pairs of participants  and developed a machine learning classifier to
automatically determine whether the messages of an author were UpSpeak (directed towards a person of higher status) or DownSpeak (directed towards a person of lower status). 
With a slightly different formulation of the task, \namecite{Gilbert:2012:PSW:2145204.2145359} used logistic regression to classify power relationships in the Enron corpus and identified the most predictive
phrases. 
Besides formulating the task as a classification task, ranking approaches
have been explored as well  \cite{Diehl:2007:RIS:1619645.1619733,Nguyen:2014vn,I13-1042}. 
For example, \namecite{I13-1042} predicted the ranking of participants in political debates
according to their relative poll standings.

Studies based on external power structures, such as the organizational structure of a company, treat  power relations as static. 
Recent studies have adopted more dynamic notions of power. 
For example, \namecite{C12-1138} discuss a
setting with an employee in a Human Resources department
who interacts with an office manager. The HR employee has power over the office manager 
when the situation is about enforcing a HR policy, but the power relation will be reversed when 
the topic is allocation of new office space.
In their study using the Enron corpus, they compared manual annotations of situational power
with the organization hierarchy and found that these were not well aligned.
Other studies have focused on a more dynamic view of power as arising through asymmetries with respect to needed resources or other goals, as characterized in consent-based theories of power such as exchange theory \cite{socialpsypower2010}.  This would include such investigations as identifying persons who are pursuing power \cite{6337148,swayamdipta2012pursuit} and detecting influencers \cite{W12-2105,HCRE:HCRE1390,Nguyen:2014vn,Quercia2011}.  This could also include studying how language use changes when users change their status in online communities  \cite{danescu2012echoes}.

Depending on the conceptualization of power and the used dataset, labels for the relations or roles of individuals have been
 collected in different ways, such as based on the organizational structure of Enron \cite{bramsen2011extracting,Gilbert:2012:PSW:2145204.2145359},  the number of followers in Twitter \cite{Danescu-Niculescu-Mizil:2011:MMW:1963405.1963509},
 standings in state and national polls to study power in political debates \cite{I13-1042},
 admins and non-admins in Wikipedia \cite{W11-0707,danescu2012echoes},
and manual annotation  \cite{W12-2105,Nguyen:2014vn,prabhakaran2013written}.

Many computational approaches within this sphere  build on a foundation from pragmatics related to speech act theory \cite{austin1975things,searle1969speech}, which has most commonly been represented in what are typically referred to as conversation, dialog or social acts \cite{W11-0707,E12-1079}.
Such categories can also be combined into sequences \cite{6337148}.
Other specialized representations are also used, such as features related to turn taking style \cite{I13-1042,swayamdipta2012pursuit},  topic control \cite{Nguyen:2014vn,D14-1157,C12-1155},
and `overt displays of power', which \namecite{Prabhakaran:2012:POD:2382029.2382104} define
as utterances that constrain the addressee's actions beyond what the underlying
dialog act imposes.

\paragraph*{Politeness}  Polite behavior contributes to maintaining social harmony and avoiding social conflict \cite{holmes2013introduction}. 
Automatic classifiers to detect politeness have been developed to study politeness strategies
on a large scale. 
According to politeness theory by \namecite{brown1987politeness}, three social factors
influence linguistically polite behavior: social distance, relative power, and ranking of the imposition (i.e., cost of the request).
Drawing from this theory, \namecite{W11-0711} performed a study on the Enron corpus by
training classifiers to automatically detect formality and requests. Emails that contained requests
or that were sent to people  of higher ranks indeed tended to be more formal.
According to politeness theory, speakers with greater power than their addressees are
expected to be less polite \cite{brown1987politeness}.
\namecite{P13-1025} developed a politeness classifier and found that  in Wikipedia
polite editors were  more likely to achieve higher status, but
once promoted, they indeed became less polite. 
In StackExchange, a site with an explicit reputation system,
users with a higher reputation were less polite than
users with a lower reputation.
Their study also revealed new interactions between
politeness markings (e.g., `\emph{please}') and morphosyntactic context.

\subsection{Style Shifting}
\label{style_shifting}
According to \namecite{labov1972}, there are no single-style speakers since speakers may switch between styles (style-shifting) depending on their communication partners (e.g., addressee's age, gender and social background).
Besides the addressee, other factors such as the topic (e.g., politics vs. religion)
or the context (e.g., a courtroom vs. family dinner) can contribute to style shifting.
In early studies, Labov stated that "\textit{styles can be arranged along a single dimension, measured by the amount of attention paid to speech}" \cite{labov1972}, which thus views style shifting as mainly something responsive.
The work by Labov on style has been highly influential, but not everyone agreed with his explanation
for different speech styles.
We will discuss two theories (Communication Accommodation Theory and Audience Design) that have received much attention in both sociolinguistics and computational linguistics and that focus on the role of audiences and addressees on style.
Even more recent theories are emphasizing the agency of speakers as they employ different styles to represent themselves in a certain way or initiate a change in
the situation. Besides switching between styles, multilingual speakers may also switch between languages or dialects.  This is discussed in more depth in Section \ref{sec:multilingualism}.

\paragraph*{Communication Accommodation Theory}  
Communication Accommodation Theory (CAT) \cite{LSY:2923744,giles19911,soliz2014relational}  seeks
to explain why speakers accommodate\footnote{The  phenomenon of
adapting to the conversation partner has also been known as `alignment', `coordination' and `entrainment'.} to each other during conversations.
Speakers can shift their behavior to become more similar (convergence) or more different (divergence) to their conversation partners.
Convergence reduces the social distance between speakers and  converging speakers are often viewed as more favorable and cooperative.
CAT has been developed in the 1970s and has its roots in the field of social psychology. 
While CAT has been studied extensively in controlled settings, e.g., \namecite{Gonzales01022010}, only recently studies have been performed in uncontrolled settings such as  Twitter conversations \cite{Danescu-Niculescu-Mizil:2011:MMW:1963405.1963509}, online forums \cite{C14-1044}, Wikipedia Talk pages and Supreme Court arguments \cite{danescu2012echoes}, and even movie scripts \cite{Danescu-Niculescu-Mizil:2011:CIC:2021096.2021105}. 

Speakers accommodate to each other on a variety of dimensions, ranging from pitch and gestures, to the words that are used.
Within computational linguistics, researchers have focused on measuring linguistic accommodation. LIWC has frequently been employed in these studies to capture stylistic accommodation, for example as reflected in the use of pronouns 
\cite{Danescu-Niculescu-Mizil:2011:CIC:2021096.2021105,Danescu-Niculescu-Mizil:2011:MMW:1963405.1963509,C14-1044,Niederhoffer01122002}. Speakers do not necessarily converge on all dimensions \cite{giles19911}, which has also been observed on Twitter \cite{Danescu-Niculescu-Mizil:2011:MMW:1963405.1963509}.
Although earlier studies  used correlations of specific features between participants,
on turn-level or overall conversation-level \cite{P11-2020,Niederhoffer01122002,Scissors:2009:CWT:1518701.1518783}, these correlations fail to capture the temporal aspect of accommodation. 
The measure developed by \namecite{Danescu-Niculescu-Mizil:2011:MMW:1963405.1963509} is based on the increase in probability of a response
containing a certain stylistic dimension  given that the original message contains that specific stylistic dimension. \namecite{wang2014} used a measure based on repetition of words (or syntactic structures) between  target and prime posts.
\namecite{C14-1044} proposed a measure that takes into account that speakers differ in their tendency to accommodate to others.
Similarly, \namecite{jain2012unsupervised} used a Dynamic Bayesian Model to induce latent style states that group related style choices together in a way that reflects relevant styles within a corpus. They also introduce global accommodation states that provide more context in identification of style shifts in  interactions that extend for more than a couple of turns.  

Social roles and orientations taken up by speakers influence how conversations play out over time
and  computational approaches to measure accommodation have been used to study power dynamics \cite{Danescu-Niculescu-Mizil:2011:MMW:1963405.1963509,danescu2012echoes,C14-1044}. 
In a study on power dynamics in Wikipedia Talk pages and Supreme court debates, \namecite{danescu2012echoes}  found that people with a lower status accommodated more
than people with a higher status. In addition, users accommodated less once they became an admin in Wikipedia. Using the same Wikipedia data, \namecite{noble-fernandez:2015:CMCL} found that users accommodated more towards users that occupied a more central position, based on eigenvector and betweenness centrality, in the social network. Furthermore, whether a user was an admin did not had a significant effect on the amount of coordination that highly central users received.
From a different angle, \namecite{Gweon2013} studied  transactive exchange in debate contexts. Transactivity is a property of an assertion that requires that it displays reasoning (e.g., a causal mechanism) and refers to or integrates an idea expressed earlier in the discussion.
In this context, high concentrations of transactivity reflect a balance of power in a discussion. In their data, higher levels of speech style accommodation were correlated with higher levels of transactivity.

\paragraph*{Audience Design} 
In a classical study set in New Zealand, Allan Bell found that newsreaders used different styles
depending on which radio station they were talking for, even when they were reporting the same news on the same day. Bell's audience design framework \cite{LSY:2990984} 
explains style shifting as a response to audiences and shares similarities with CAT.
One of the differences with CAT is that different types of audiences are defined from the perspective of the speaker (ranging from addressee to eavesdropper) and thus
can also be applied to settings
in which there is only a one-way interaction (such as broadcasting).
Social media provides an interesting setting to study how audiences influence style.
In many social media platforms, such as Twitter or Facebook, multiple audiences (e.g., friends, colleagues) are collapsed into a single context. Users of such platforms often  imagine an audience when writing messages and they may target messages to different audiences 
 \cite{marwick2011tweet}.

Twitter has been the focus of several recent large-scale studies on audience design.
In a study on how audiences influence the use of minority languages on Twitter,
\namecite{nguyenicwsm2015} showed how  characteristics of the audience influence 
language choice on Twitter by analyzing tweets from multilingual users in the Netherlands using
automatic language identification.
Tweets directed to larger audiences were more often written in Dutch, while within conversations
users often switched to the minority language. In another study on audience on Twitter,
\namecite{bamman2015} showed that incorporating features of the audience improved sarcasm detection.
Furthermore, their results suggested that users tend to use the hashtag \#sarcasm when they are less familiar
with their audience. 
\namecite{pavalanathan2015linguistic} studied two types of non-standard lexical variables: those strongly
associated with specific geographical regions of the United States and variables that were frequently used in Twitter but considered
non-standard in other media. The use of non-standard lexical variables was higher in messages with user mentions, which are usually intended for smaller audiences,
and lower in messages with hashtags, which
are usually intended for larger audiences. Furthermore, non-standard lexical variables were more often used in tweets addressed to individuals from the same metropolitan area.
Using a different data source, \namecite{ICWSM148046} showed that reviewers on the TripAdvisor site adjust their style to the style of preceding reviews.
Moreover, the extent to which reviewers are influenced correlates
with attributes such as experience of the reviewer and their sentiment towards the reviewed attraction.

\subsection{Community Dynamics}
\label{community_dynamics}
As we just discussed, people adapt their language use towards their conversation partner.
Within communities,  norms emerge over time through interaction  between members, such
as the use of slang words and domain-specific jargon \cite{Danescu-Niculescu-Mizil:2013:NCO:2488388.2488416,Nguyen:2011:LUR:2021109.2021119}, or conventions for indicating retweets in Twitter \cite{ICWSM124661}.
Community members employ such markers to signal their affiliation.
In an online gangs forum, for example,
graffiti style features were used to signal  group affiliation \cite{E14-1012}.
To become a core member of a community, members adopt such community norms.
As a result, often a change in behavior can be observed when someone joins a community.
Multiple studies have reported that members of online communities decrease their use of
first person singular pronouns (e.g., `\textit{I}') over time and increase their use of first person plural pronouns (e.g., `\textit{we}') \cite{JCC4:JCC402,Danescu-Niculescu-Mizil:2013:NCO:2488388.2488416,Nguyen:2011:LUR:2021109.2021119},
suggesting a stronger focus on the community. Depending on the frequency of use and social factors, local accommodation effects could influence how languages change in the long term  \cite{labov2010a,labovsocial}.
Fine-grained, large-scale analyses of language change are difficult in offline settings, but the emergence of online communities
has enabled computational approaches for analyzing language change within communities.

Early investigations of this topic
were based on data from non-public communities, such as email exchanges between students during a course \cite{HCRE:HCRE341} and data from the Junior Summit ’98, 
an  online community where children from across the world discussed global issues  \cite{JCC4:JCC402,huffaker2006computational}.
In these communities, members joined at the same time. Furthermore, the studies
were based on data spanning only several months.

More recent studies have used data from  public, online communities, such
as online forums and review sites.  Data from
these communities typically span longer time periods (e.g., multiple years).
Members join these communities intermittently
and thus, when new users join, community norms have already been established.
\namecite{Nguyen:2011:LUR:2021109.2021119} analyzed an online breast cancer community, in which
long-time members 
 used forum-specific jargon,
highly informal style, and showed familiarity and emotional involvement with other members.
Time periods were represented
by the distribution of high frequency words and measures such as Kullback-Leibler divergence
were used to study how language changed over time.
Members who joined the community showed increasing conformity to community norms during
the first year of their participation. Based on these observations, a 
model was developed to determine membership duration.
\namecite{hemphill2012learning} also studied how members adopt community norms over time
but focused specifically on gender differences. They studied
changes in the use of various characteristics, such as  hedging, word/sentence complexity and vocabulary richness,
in IMDb (the Internet Movie Database),
a community in which males tend to receive higher prestige than females.

Not only members change their behavior over time as they participate in a community,
communities themselves are also constantly evolving.
\namecite{Kershaw:2016:TML:2835776.2835784} identified and analyzed word innovations in Twitter and Reddit based on variation in frequency, form and meaning. They performed their analyses on a global level, i.e., the whole dataset, and on a community level, based on applying a community detection algorithm to the Reddit data and grouping the geotagged tweets by geopolitical units. 

Language change
on both member-level and community-level was analyzed by
\namecite{Danescu-Niculescu-Mizil:2013:NCO:2488388.2488416} in two beer review communities.
Language models were created based on monthly snapshots to capture
the linguistic state of a community over time. Cross-entropy was then used
to measure how much a certain post deviated from a language model.
Members in these communities turned out to follow  a two-stage lifecycle: 
They first align with the language of the community (innovative learning phase),
however at some point they stop adapting their language (conservative phase).
The point at which members enter the conservative phase turned out to be 
dependent on how long a user would end up staying in the community.

These studies illustrate the potential of using large amounts of online data
to study language change in communities in a quantitative manner. However, in such analyses biases in the data should be considered carefully, especially when the dynamics and content of the data are not understood fully. For example, \namecite{10.1371/journal.pone.0137041} call into question the findings on linguistic change based on the Google books corpus, due to its bias towards scientific publications. Furthermore, they point out that prolific  authors in the dataset can influence the findings as well.

\section{Multilingualism and Social Interaction}
\label{sec:multilingualism}
Languages evolve due to the interaction of speakers within and outside their speech communities. Within sociolinguistics, multilingual speakers and  speech communities have been studied widely with respect to the contexts and conditions of language mixing and/or switching across languages. We  use the term `multilingual speaker' for someone who has a repertoire of various languages and/or dialects and who may mix them depending on contextual factors like occasion (e.g., home vs. work) and conversation partners (e.g., family vs. formal encounters). This section is dedicated to computational approaches for analyzing multilingual communication in relation to the social and linguistic contexts. We first start with a brief introduction into multilingual communication from a sociolinguistic point of view. Later, we expand the discussion to include the analysis of multilingual communication using computational approaches.

Human mobility is one of the main reasons for interaction among speakers of different languages. \namecite{weinreich1953languages} was one of the first to explain why and how languages come into contact and evolve under each other's influence in a systematic manner. 
Sociolinguists \cite{auer1988conversation,gumperz1982discourse,myers2002contact,poplack1988social} have studied various aspects of language contact and mixing across different contact settings.

Language mixing and code-switching are used interchangeably  and there is not always a consensus on the terminology.
According to \namecite{gumperz1982discourse}, language mixing refers to  the mixing of languages within the same text or conversation.  
 \namecite{liwei1998codeswitching} describes language alternations at or above the clause level and calls it code-mixing. \namecite{romaine1995bilingualism} differentiates between inter-sentential (i.e., across sentences) and intra-sentential (i.e., within the same sentence) switches. \namecite{poplack1988social} refer to complete languages shifts of individual users as code-switching. 
  
 Language mixing spans across a continuum ranging from occasional switches (e.g., words or fixed multi-word expressions) to more structural ones (e.g., morphological, syntactic borrowings). The duration and intensity of interaction between 
   speakers of contact languages influence the types of switches.
 When the frequency of switched words increases in use, they may get established in the speech community and become borrowed/loan words  (e.g., hip hop-related  
Anglicisms in a German hip hop forum \cite{garley2012beefmoves}). 

Earlier studies on language mixing were mostly based on multilingual spoken data collected in controlled or naturalistic settings \cite{auer1988conversation,scotton1995social}. Nowadays, the wide-spread use of internet in multilingual populations provides ample opportunities for large-scale and in-depth analyses of mixed language use in online media \cite{danet2007multilingual,hinnenkamp2008deutsch,hinrichs2006codeswitching,paolillo2001language,tsaliki2003globalization}. Still most of these studies focus on qualitative analyses of multilingual online communication with limited data in terms of size and duration.

%
%

The rest of this section presents  a discussion of data sources for studying multilingual communication on a large scale (Section \ref{sec:multi_data}).
Consequently, we discuss research on adapting various NLP tools to process mixed-language texts  (Section \ref{sec:multi_processing}).
We conclude this section with a discussion of studies that analyze, or even try to predict, the use of multiple languages
in multilingual communication (Section \ref{sec:multi_analysis}).

\subsection{Data Sources}
\label{sec:multi_data}
In sociolinguistics, conversational data is usually collected by the researchers themselves, either among small groups of speakers at different times \cite{seza2007,BIL:3107556} or from the same group of speakers longitudinally \cite{milroy1987,Trudgill2003}. The manual transcription and annotation of data is time-intensive and costly. Multilingual data from online environments is usually extracted in 
small volumes and for short periods. Automatic analysis of this type of data has been difficult for most languages, especially when resources or technical support are lacking.

Within computational linguistics, there is a growing interest in the automatic processing of mixed-language texts.
 \namecite{lui2014automatic} and \namecite{yamaguchi2012text} studied automatic language identification in mixed-language documents from Wikipedia by artificially concatenating texts from monolingual sources into multilingual documents. However, such approaches lead to artificial language boundaries. 
More recently, social media (such as Facebook \cite{vyas2014pos}, Twitter \cite{jurgens2014twitter,penglearning,solorio2014overview} and online forums \cite{nguyen2013}) provide large volumes of data for analyzing multilingual communication in social interaction. 
Transcriptions of conversations have been explored by \namecite{solorio2008part}, however their data was limited to three speakers.
Language pairs that  have been studied for multilingual communication  include English-Hindi \cite{vyas2014pos}, Spanish-English \cite{penglearning,solorio2008learning,solorio2008part}, Turkish-Dutch \cite{nguyen2013}, Mandarin-English \cite{adelcombination,penglearning},
and French-English \cite{jurgens2014twitter}.
Besides being a valuable resource for studies on multilingual social interaction, multilingual texts
in social media have also been used to improve general purpose machine translation systems \cite{E14-1001,ling2013microblogs}.

Processing and analyzing mixed-language data often requires identification of languages at the word level. Language identification is a well-researched problem in CL and we discussed it in the context of dialect identification in Section \ref{identity_regional}. Here, we discuss language identification for mixed-language texts.
 Several datasets are publicly available to stimulate research on language identification in mixed-language texts, including data from the shared task on  Language Identification in Code-Switched Data  \cite{solorio2014overview} covering four different language pairs on Twitter,
romanized Algerian Arabic and French  texts from the comments section of an online Algerian newspaper \cite{cotterel2014},
 Turkish-Dutch forum posts \cite{nguyen2013} and web documents in different languages \cite{king2013labeling}.

Annotation on a fine-grained level such as individual words has introduced new challenges in the construction of datasets.
More fine-grained  annotations require more effort 
and sometimes the segments  are so short that they can no longer be clearly attributed to a particular language.
For example, annotating the language of named entities remains a challenge in mixed-language texts. Named entities have been labeled according to the context \cite{king2013labeling}, ignored in the evaluation
\cite{elfardy2012token,nguyen2013} or treated as a separate  category \cite{elfardy2012simplified,solorio2014overview}.
Annotation at  sentence-level is also challenging. For example, \namecite{zaidan2013arabic} annotated a large corpus for Arabic dialect identification using crowdsourcing. Their analysis indicated that many annotators over-identify their native dialect (i.e., they were biased towards labeling texts as written in their own dialect). \namecite{elfardy2012simplified}  presented guidelines
to annotate texts written in dialectal variants of Arabic and Modern Standard Arabic  on a word level.

\subsection{NLP Tools for Multilingual Data}
\label{sec:multi_processing}
Most of the current NLP tools, such as parsers, are developed for texts written in a single language. Therefore, such tools are not optimized for processing texts containing multiple languages.
In this section, we discuss the development of NLP tools that specifically aim to support the  processing of multilingual texts.
We start with research on automatic language identification, which is  an important step in the preprocessing pipeline of many
language-specific  analysis tasks. Mixed-language documents have introduced
new challenges to this task.
We then continue with a discussion of work on various other NLP tools (e.g., parsers, topic modeling).

\paragraph*{Automatic Language Identification}
Automatic language identification is often the first step for systems that process mixed-language texts \cite{vyas2014pos}.
Furthermore, it  supports large-scale analyses of patterns in multilingual communication \cite{jurgens2014twitter,kim2014sociolinguistic,papalexakis2014predicting}.
Most of the earlier research on automatic language identification focused on document-level identification of a single language \cite{baldwin2010language}.
To handle mixed-language texts, more fine-grained approaches have been explored, ranging from language identification at the sentence \cite{elfardy2013sentence,zaidan2013arabic,zampieri2014report} and word level \cite{elfardy2012token,king2013labeling,nguyen2013,solorio2014overview,L14-1086},  approaches for  text segmentation \cite{yamaguchi2012text},
and estimating the proportion of the various languages used within documents \cite{lui2014automatic,prager1999linguini}.
Depending on the application, different approaches may be suitable, but studies that analyze patterns in multilingual communication have mostly focused on word-level identification
\cite{nguyen2013,solorio2014overview}.
Off-the-shelf tools developed for language identification at the document-level (e.g., the TextCat program  \cite{cavnar1994n})
are not effective for word-level identification \cite{alex2005unsupervised,nguyen2013}.
Language models \cite{elfardy2012token,nguyen2013} and dictionaries \cite{alex2005unsupervised,elfardy2012token,nguyen2013},
which are also commonly used in automatic language identification at the document level, have been explored.
Furthermore, the context around the words has been exploited using Conditional Random Fields to improve performance on
language identification at the word level  \cite{king2013labeling,nguyen2013}.

\paragraph*{Parsing}
Early studies on language mixing within  computational linguistics
focused on developing grammars to model language mixing (e.g., \namecite{joshi1982processing}). 
However, the models developed in these early studies were not tested on empirical data.
The more recently developed systems have been validated on large, real-world data.
\namecite{solorio2008part} explored various strategies to combine monolingual taggers to parse mixed-language texts.
The best performance was obtained by including the output of the monolingual parsers as features in a machine learning algorithm. 
\namecite{vyas2014pos} studied the impact of different preprocessing steps on POS tagging
of English-Hindi data collected from Facebook.
Language identification and
transliteration were the major challenges that impacted  POS performance.

\paragraph*{Language and Topic Models}
Language models have  been developed to improve speech recognition for mixed-language speech, by adding POS and language information to the language models   \cite{adelcombination}
or by incorporating syntactic inversion constraints \cite{C12-1102}.
\namecite{penglearning} developed a topic model that infers language-specific topic distributions
based on mixed-language  text. The main challenge for their model was aligning the inferred topics across languages.

\subsection{Analysis and Prediction of Multilingual Communication}
\label{sec:multi_analysis}
According to \namecite{thomason2001language}, \namecite{gardner-chloros2004assumptions}, and \namecite{bhattbolonyai2011code},
social factors (e.g., attitudes and motives of the speakers, social and political context) are as important as linguistic factors in multilingual settings. 
Large-scale analysis of social factors  in multilingual communication has only recently been possible with
the availability of automatic language identification tools.

Twitter is frequently used as a resource for such studies.
Focusing on language choice at the user level, researchers have extracted network structures,
based on followers and followees \cite{Eleta2014424,kim2014sociolinguistic}, or mentions and retweets \cite{hale2014global},
and analyzed the relation between the composition of such networks and the language choices of users.
Users tweeting in multiple languages are often found to function as a bridge between communities 
tweeting in one language. 
Besides analyzing language choice at the user level, there is also an interest in the language choices for individual tweets.  
\namecite{jurgens2014twitter} studied tweets written in one language but containing hashtags in another language. 
Automatic language identification was used to identify  the languages of the tweets. However, as they note, some  tweets  were written in another language because they were automatically generated by applications
rather than being a conscious choice of the user.
 \namecite{nguyenicwsm2015} studied users in the Netherlands who tweeted in 
a minority language (Limburgish or Frisian) as well as in Dutch. Most tweets were written in Dutch, but during conversations users often switched
to the minority language (i.e., Limburgish or Frisian). \namecite{10.1371/journal.pone.0061981} analyzed the geographic distribution of languages in multilingual regions and cities (such as New York and Montreal) using Twitter. 

In addition to the analysis of patterns in multilingual communication, several studies have explored the automatic prediction
of language switches. The task may seem similar to automatic language identification, yet there are differences between the two tasks.
 Rather than determining the language of an utterance,
it 
 involves predicting whether the language of the next utterance  is the same \textit{without} having access to the next utterance itself. 
\namecite{solorio2008learning} were the first to predict whether a speaker will switch to another language  in English-Spanish bilingual spoken conversations based on  lexical and syntactic  features. The approach was evaluated using standard machine learning metrics as well
as  human evaluators who rated the naturalness/human-likeness of the sentences the system generated. 
 \namecite{papalexakis2014predicting} predicted when multilingual users switch between languages  in a Turkish-Dutch online forum
using various features, including features based on multi-word units and emoticons.
%
%

\section{Research Agenda}
 \label{sec:conclusion}
 Computational sociolinguistics is an emerging multidisciplinary field.
Closer collaboration between sociolinguists and computational linguists 
could be beneficial to researchers from both fields.
In this article, we have outlined some challenges related
to differences in data and methods that must be addressed in order for synergy to be effective. In this section, we summarize the main challenges for advancing the field of computational sociolinguistics.  These fall under three main headings, namely, expanding the scope of inquiry of the field, adapting  methods to increase compatibility, and offering tools.

\subsection{Expanding the Scope of Inquiry}
The field of computational linguistics has begun to investigate issues that overlap with those of the field of sociolinguistics. The emerging availability of data that is of interest to both communities is an important factor, but in order for real synergy to come out of this, additional angles in the research agendas and tuning of the methodological frameworks in the respective communities would be needed.

\paragraph*{Going beyond lexical and stylistic variation} 
Many studies within CL
focus on lexical variation (e.g., Section~\ref{sec:identity}  on social identity), possibly driven by the focus on prediction tasks.
Stylistic variation has also received attention. Several of the discussed studies
focus on variation in the usage of functional categories.  For example, they zoom in on  the usage of determiners, prepositions and pronouns 
for studying linguistic style accommodation (in Section \ref{style_shifting}).
Others  employ measures such as average word and sentence length (e.g., in Section \ref{sec:identity}). Advances in the area of stylometry \cite{ASI:ASI21001} 
 could inspire the exploration of more fine-grained features to capture style. 
 Besides lexical and stylistic variation, linguistic variation also occurs on many other levels. Some computational studies have focused on phonological \cite{eisensteinphonological,jain2012unsupervised,anna2015} and syntactic \cite{doyle2014mapping,gianfortoni2011modeling,johannsen-hovy-sogaard:2015:CoNLL,Wiersma11102010} variation, but so far the number of studies is limited. In combination with the surge in availability of relevant data, these examples suggest that there seems to be ample opportunities for an  extended scope. 

\paragraph*{Extending  focus to other social variables}
A large body of work exists on the modeling of gender, age and regional variation (Cf. Section~\ref{sec:identity}).
Other variables, like social class \cite{labov1966},
have barely received any attention so far within computational sociolinguistics.
Although it is more difficult to obtain labels for some social variables, they are essential for a richer understanding of language variation and more robust analyses.

\paragraph*{Going beyond English and monolingual data} 
The world is multilingual and multicultural, but English has received much more attention within computational
sociolinguistics than other languages. There is a need for research to validate the generalizability of  findings based on English data for other languages \cite{danet2007multilingual}. 
Furthermore, most studies within computational linguistics generally assume that
texts are written in one language. However, these assumptions may not hold, especially in social media. A single user may use multiple languages, sometimes even within a syntactic unit, while most 
NLP tools are not optimized to process such texts.
Tools that are able to process mixed-language texts will support  the analysis of  such data and
shed more light on the social and linguistic factors involved in multilingual communication.

\paragraph*{From monomodal to multimodal data}  Another recommendable shift in scope would be a stronger focus on multimedia data. Video and audio recordings with a speech track encapsulate a form of language in which the  verbal and nonverbal dimensions of human communication are available in an integrated manner and they represent a  rich source for the study of social behavior. Among the so-called paralinguistic aspects for which detection models and evaluation frameworks exist are age, gender and affect \cite{schuller2010interspeech}.
The increasing volumes of recordings of spoken dialogue and aligned transcriptions, e.g.,  in  oral history collections \cite{oralhistory2013,jong2014}, meeting recording archives \cite{1198793}, and video blogs \cite{Biel:2013:HYP:2522848.2522877}, can add new angles to the investigation of sociolinguistic variation. In particular, the study of the interaction between  (transcribed) speech, non-speech (laughter, sighs, etc.), facial expression and gestures is a promising area for capturing and predicting social variables as well as the related affective layers.

\subsection{Adapting Methodological Frameworks to Increase Compatibility}

To make use of the rich repertoire of theory and practice from sociolinguistics and to contribute to it, we have to appreciate the methodologies that underlie sociolinguistic research, e.g., the \emph{rules of engagement} for joining into the ongoing scientific discourse.  However, as we have highlighted in the methodology discussion earlier in the article,  the differences in values between the communities  can be perceived as a divide. While the CL community has experienced a history in which theory and empiricism are treated as the extreme ends of a spectrum, in the social sciences there is no such dichotomy, and empiricism contributes substantially to theory. 
Moving forward, research within computational sociolinguistics should build on and seek to partner in extending existing sociolinguistic theories and insights.  This requires  placing a strong focus on the interpretability of the developed models.
 The feasibility of such a shift in attention can be seen when observing successes of applied computational sociolinguistics work that has been adopted in other fields like health communication \cite{mayfield2014} and education \cite{discanal1}.

\paragraph*{Controlling for multiple variables}
Sociolinguistic studies typically control for multiple social variables (e.g., gender, age, social class, ethnicity).
However, many studies in computational sociolinguistics focus on individual variables (e.g., only gender, or only age),
which can be explained by the focus on social media data. The uncontrolled nature of social media
makes it  challenging  to obtain data about the
social backgrounds of the speakers
and to understand the various biases that such datasets might have.  The result is that models are frequently confounded, which results in low interpretability as well as limited justification for generalization to other domains.

On the other hand, much work in the CL community has focused on structured modeling approaches that take a step towards addressing these issues \cite{joshi2012multi,joshi2013s}.  These approaches are very similar to the hierarchical modeling approaches used in sociolinguistic research to control for multiple sources of variation and thus avoid misattributing weight to extraneous variables.  A stronger partnership within the field of CL between researchers interested in computational sociolinguistics and researchers interested in multi-domain learning would be valuable for addressing some of the limitations mentioned above. In this regard, inferring demographic variables automatically (see Section~\ref{sec:identity}) may also
help, since  predicted demographic variables could  be used in structuring the models. Another approach is the use of census data when location data is already available. For example, \namecite{10.1371/journal.pone.0113114} studied  lexical change in social media by using census data to obtain demographic information for the geographical locations. They justified their approach by assuming that lexical change is influenced by the demographics of the population in these locations, and not necessarily by the demographics of the particular Twitter users in these locations. 

\paragraph*{Developing models that generalize across domains}
Many of the studies within the area of computational sociolinguistics have focused on a single domain. However, domain effects can influence the findings, such as which
features are predictive for gender (e.g., \namecite{herring2006gender}). 
Studies considering  multiple domains enable
distinguishing  variables that work differently in different contexts, and therefore
improve the interpretation of the findings.
Recently, several studies  within the area of computational sociolinguistics
have performed experiments across domains \cite{sap2014,sarawgi2011gender} and explored the effectiveness of domain adaptation approaches \cite{nguyen2011author,E14-1012}.
Another approach involves reconsidering the features used in an attempt to include more features with a deep connection with the predicted variable of interest.  For example, \namecite{gianfortoni2011modeling} show that  features such
as n-grams, usually reported to be predictive  for gender classification, did not perform well after controlling for occupation in a blog corpus, but pattern-based features inspired by findings related to gender-based language practices did.

\paragraph*{Using sociolinguistics and the social sciences as a source of inspiration for methodological reflection}
Going forward, we need to appreciate where our work stands along an important continuum that represents a fundamental tension in the social sciences: qualitative approaches that seek to preserve the complexity of the phenomena of interest, versus quantitative approaches that discretize (but thereby also simplify) the phenomena to achieve more generalizability. For computational linguistics, a primarily quantitative field,  work from research areas with a less strong or less exclusive
focus on quantitative measures, such as sociolinguistics and the social sciences,
could serve as a source of inspiration for methodological reflection.
In this survey, we have questioned the operationalizations of
the concepts of gender (Section~\ref{sec:gender}), age (Section~\ref{sec:age}) and language variety (Section~\ref{sec:geo}) as discrete and static categories, based on insights from sociolinguistics.
More critical  reflection on such operationalizations could lead to 
a deeper insight into the limitations of the developed models and the incorrect predictions that they sometimes make.

\subsection{Tuning NLP Tools to Requirements of Sociolinguistics Research}

As a final important direction, we should consider what would be required for NLP tools to be supportive
for sociolinguistic work.

\paragraph*{Developing models that can guide users of data analysis systems in taking next steps}
Sociolinguists are primarily interested in new insights about language use. In contrast,
much of the work within CL is centered around highly specific analysis tasks that are isolated from scenarios of use,
and the focus on the obtained performance figures  for such tasks is fairly dominant. As \namecite{manning2016computational} mentions: "\textit{[..], there has been an over-focus on numbers, on beating the state of the art}".
Only for few analysis methods, validation of the outcomes has been pursued  (e.g., have we measured the right thing?)
in view of the potential for integration of the models outside lab-like environments. 
Furthermore many of the models developed within CL make use of thousands of features.
As a result, their value for practical data exploration tasks is therefore often limited.
Sparse models, such as used in \namecite{eisenstein2011discovering}, that identify small sets of predictive features would be more suited for exploratory analysis.   However, when the focus is on interpretability of the models, we must consider that the resulting average prediction performance of interpretable models may be lower \cite{E14-1012}.

\paragraph*{Developing pre-processing tools to support the analysis of language variation}
The performance of many developed  NLP tools is lower on informal text. 
For example, POS taggers
perform less well on texts written by certain user groups (e.g., younger people \cite{hovyage2015}) or on texts in certain language
varieties (e.g., African American Vernacular English \cite{anna2015}).
One of the approaches to improve the performance of tools has been to normalize the texts, but as \namecite{eisenstein2013bad} argues,
doing so is removing the variation that is central to the study of sociolinguistics.
To support deeper sociolinguistic analyses and to go beyond shallow features, we thus need pre-processing tools,
such as POS taggers, that are able to handle the variation found in informal texts and that are not biased
towards certain social groups.

\section{Conclusion}
While the computational linguistics field has historically emphasized interpretation and manipulation of the propositional content of language, another valid perspective on language is that it is a dynamic, social entity.  While some aspects of language viewed from a social perspective are predictable, and thus behave much like other aspects more commonly the target of inquiry in the field, we must acknowledge that linguistic agency is a big part of how language is used to construct social identities, to build and maintain social relationships, and even to define the boundaries of communities.  
The increasing research on social media data has contributed to the insight that text can be considered as a data source that captures multiple aspects and layers of human and social behavior. The recent focus on text as social data and the emergence of computational
social science are likely to increase the interest within the computational linguistics community on sociolinguistic topics.
In this article, we have defined and set out a research agenda for the emerging field of
`\textit{Computational Sociolinguistics}'.  We have aimed
to provide a comprehensive overview of studies published within the field of CL
that touch upon sociolinguistic themes in order to provide an overview of what has been accomplished so far and where there is room for growth. In particular, we have endeavored to illustrate how the large-scale data-driven methods
of our community can complement existing sociolinguistic studies, but also
how sociolinguistics can inform and challenge
our methods and assumptions.

\starttwocolumn
 \begin{acknowledgments}
Thanks to Mari\"{e}t Theune, Dirk Hovy and Marcos Zampieri for helpful comments
on the draft.
Thanks also to the anonymous reviewers for their valuable and detailed feedback.
This work was funded in part through NSF grant ACI-1443068 and ARL grant W911NF-11-2-0042.
The first author was supported by the Netherlands Organization for Scientific Research (NWO) grant 640.005.002 (CATCH project FACT).
The second author was supported by the Digital Humanities Research Grant from Tilburg University and a fellowship from the Netherlands Institute of Advanced Study in Humanities and Social Sciences.     
\end{acknowledgments}

\bibliographystyle{fullname}
\bibliography{bib}
\end{document}